\documentclass[runningheads]{llncs}
\usepackage[T1]{fontenc}
\usepackage{graphicx}
\usepackage{soul}
\usepackage{url}
\usepackage[hidelinks]{hyperref}
\usepackage[utf8]{inputenc}
\usepackage{graphicx}
\usepackage{amsmath}
\usepackage{booktabs}
\usepackage{algorithm}
\usepackage{algorithmic}
\usepackage{multirow}
\usepackage{multicol}
\usepackage{bm}
\usepackage{underscore}
\usepackage{amsfonts}
\usepackage{xcolor}
\usepackage{color, colortbl}
\usepackage{makecell}
\usepackage{enumitem}

\usepackage{amssymb}
\usepackage{colortbl}

\usepackage[switch]{lineno}

\definecolor{mygray}{gray}{.9}
\definecolor{gray2}{gray}{.78}
\definecolor{gray3}{gray}{.7}
\definecolor{gray4}{gray}{.6}
\definecolor{gray5}{gray}{.5}
\newcommand{\CC}{\cellcolor{mygray}}

\newcommand{\rmnum}[1]{\romannumeral #1}
\newcommand{\Rmnum}[1]{\expandafter\@slowromancap\romannumeral #1@}

\begin{document}
\title{Rethinking Uncertainly Missing and Ambiguous Visual Modality in Multi-Modal \\ Entity Alignment}


\author{
Zhuo Chen\inst{1}
\and
Lingbing Guo\inst{1}
\and
 Yin Fang\inst{1}
\and
 Yichi Zhang\inst{1}
\and
Jiaoyan Chen\inst{4}
\and
Jeff Z. Pan\inst{5}
\and
Yangning Li \inst{6}
\and
Huajun Chen\inst{1,2}
\and
Wen Zhang\inst{3}\thanks{Corresponding author.}
}

\institute{College of Computer Science, Zhejiang University, Hangzhou, China
\and
Donghai laboratory, Zhoushan, China 
\and
School of Software Technology, Zhejiang University, China
\email{\{zhuo.chen,lbguo,fangyin,zhangyichi2022,zhang.wen,huajunsir\}@zju.edu.cn}
\and
The University of Manchester \& University of Oxford, UK \\
\email{jiaoyan.cheni@manchester.ac.uk}
\and
School of Informatics, The University of Edinburgh, Edinburgh, UK \\
\email{https://knowledge-representation.org/j.z.pan/}
\and
Shenzhen International Graduate School, Tsinghua University, Shenzhen, China\\
\email{liyn20@mails.tsinghua.edu.cn}
}
\authorrunning{Z. Chen et al.}
\maketitle              

\begin{abstract}
As a crucial extension of entity alignment (EA), multi-modal entity alignment (MMEA) aims to identify identical entities across disparate knowledge graphs (KGs) by exploiting associated visual information. However, existing MMEA approaches primarily concentrate on the fusion paradigm of multi-modal entity features, while neglecting the challenges presented by the pervasive phenomenon of missing and intrinsic ambiguity of visual images.
In this paper, we present a further analysis of visual modality incompleteness, benchmarking latest MMEA models on our proposed dataset {MMEA-UMVM}, where
the types of alignment KGs covering bilingual and monolingual, with standard (non-iterative) and iterative training paradigms to evaluate the model performance. 
Our research indicates that, in the face of modality incompleteness, models succumb to overfitting the modality noise, and exhibit performance oscillations or declines at high rates of missing modality. This proves that the inclusion of additional multi-modal data can sometimes adversely affect EA. 
To address these challenges,  we introduce {UMAEA}, a robust multi-modal \textbf{e}ntity \textbf{a}lignment approach designed to tackle \textbf{u}ncertainly \textbf{m}issing and \textbf{a}mbiguous visual modalities.
It consistently achieves SOTA performance across all 97 benchmark splits, significantly surpassing existing baselines with limited parameters and time consumption, while effectively alleviating the identified limitations of other models. 
Our code and benchmark data are available at {\color{blue} \url{https://github.com/zjukg/UMAEA}}.


\end{abstract}

\section{Introduction}
\begin{figure}[!htbp]
  \centering
\includegraphics[width=1.0\linewidth]{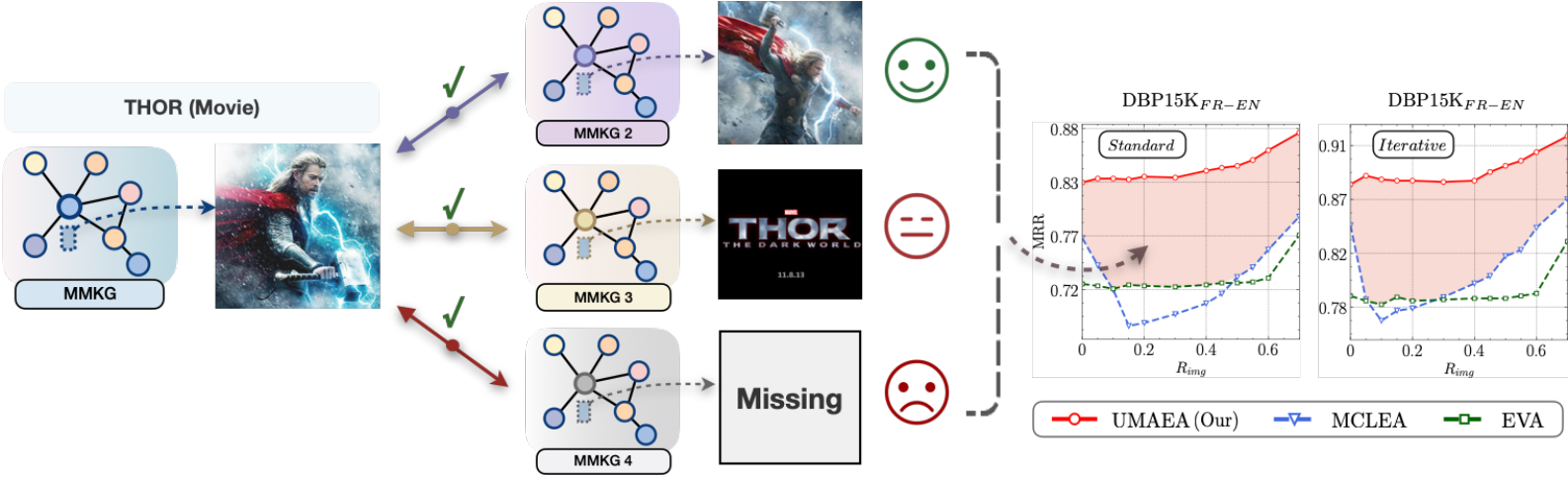}
  \caption{Phenomenon for missing and ambiguous visual modality in MMEA, where our UMAEA attains superior performance compared to MCLEA \cite{DBLP:conf/coling/LinZWSW022} and EVA \cite{DBLP:conf/aaai/0001CRC21}. 
  }
  \label{fig:Introcase}
\end{figure}
Recently entity alignment (EA) has attracted wide attention as a crucial task for aggregating knowledge graphs (KGs) from diverse data sources.
Multi-modal information, particularly visual images, serves as a vital supplement for entities. 
However, achieving visual modality completeness always proves challenging for automatically constructed KGs both on the Internet and domain-specific KGs.
 For instance, in the DBP15K datasets \cite{DBLP:conf/semweb/SunHL17} for EA, only a portion of the entities have attached images (e.g., $67.58\%$ in DBP15K$_{JA\text{-}EN}$ \cite{DBLP:conf/aaai/0001CRC21}). This incompleteness is inherent to the DBpedia KG \cite{DBLP:journals/semweb/LehmannIJJKMHMK15}, as not every entity possesses an associated image. 
Furthermore, the intrinsic ambiguity of visual images also impacts the alignment quality.
As illustrated  in Figure \ref{fig:Introcase}, the movie \emph{THOR} can be represented by a snapshot of the movie (star) poster or an image of the movie title itself. While individuals familiar with the Marvel universe can effortlessly  associate these patterns, machines struggle to discern significant visual feature association without the aid of external technologies like OCR and linking knowledge bases \cite{DBLP:conf/semweb/0007CGPYC21}, posing challenges for alignment tasks.
This phenomenon primarily arises from the abstraction of single-modal content, e.g., country-related images could be either national flags, landmarks or maps.

In this paper, we deliver an in-depth analysis of potential missing visual modality for MMEA.
To achieve this, we propose the {MMEA-UMVM} dataset, which contains seven separate datasets with a total of 97 splits, each with distinct degrees of visual modality incompleteness, and benchmark several latest MMEA models. 
 To ensure a comprehensive comparison, our dataset encompasses bilingual, monolingual, as well as normal and high-degree KG variations, with standard (non-iterative) and iterative training paradigms to evaluate the model performance.  
The robustness of the models against ambiguous images is discussed by comparing their performance under complete visual modality.

In our analysis, we identify two critical phenomena: {{(\rmnum{1})}} Models may succumb to overfitting noise during training, thereby affecting overall performance. {{(\rmnum{2})}} Models exhibit performance oscillations or even declines at high  missing modality rates, indicating that sometimes the additional multi-modal data negatively impacts EA and leads to even worse results than when no visual modality information is used.
These findings provide new insights for further exploration in this field. 
Building upon these observations, we propose our model UMAEA,  which alleviates those shortcomings of other models via introducing multi-scale modality hybrid and circularly missing modality imagination.
Experiments prove that our model can consistently achieve SOTA results across all benchmark splits with limited parameters and runtime, which supports our perspectives.

\section{Related Work}
Entity Alignment (EA) \cite{tkde/Sun,DBLP:journals/corr/abs-2305-14651} is the task of identifying equivalent entities across multiple knowledge graphs (KGs), which can facilitate knowledge integration.

\subsubsection{Typical Entity Alignment} methods mainly rely on the relational, attribute, and surface (or literal) features of KG entity for alignment. 
Specifically, symbol logic-based technologies are used \cite{DBLP:conf/semweb/Jimenez-RuizG11,DBLP:journals/pvldb/SuchanekAS11,DBLP:conf/ijcai/QiZCCXZZ21} to constrain the EA process via manually defined prior rules (e.g., logical reasoning and lexical matching).
Embedding-based methods \cite{tkde/Sun} eschew the ad-hoc heuristics of logic-based approaches, employing learned embedding space similarity measures for rapid alignment decisions.
Among these, {GNN-based EA models} \cite{DBLP:conf/emnlp/LiCHSLC19,DBLP:conf/aaai/SunW0CDZQ20,DBLP:conf/emnlp/LiuCPLC20,DBLP:conf/acl/WuLFWZ20,DBLP:conf/kdd/GaoLW0W022,DBLP:conf/semweb/WangCLSJHH22} emphasize local and global structural KG characteristics, primarily utilizing graph neural networks (GNNs) for neighborhood entity feature aggregation.
While {{translation-based EA methods}} \cite{DBLP:conf/ijcai/ZhangSHCGQ19,DBLP:conf/semweb/SunHHCGQ19,DBLP:conf/wsdm/XinSH0022,DBLP:conf/ijcai/CaiMZJ22,DBLP:journals/corr/abs-2304-04389} 
use techniques like TransE \cite{DBLP:conf/nips/BordesUGWY13} 
to capture the pairwise information from relational triples, positing that relations can be modeled as straightforward translations in the vector space. 

\subsubsection{Multi-modal Entity Alignment} (MMEA) normally leverages visual modality as supplementary information to enhance EA, with each entity accompanied by a related image.
Specifically, 
Chen et al. \cite{DBLP:conf/ksem/ChenLWXWC20} propose to combine knowledge representations from different modalities, minimizing the distance between holistic embeddings of aligned entities.
Liu et al. \cite{DBLP:conf/aaai/0001CRC21} use a learnable attention weighting scheme to assign varying importance to each modality. 
Chen et al. \cite{DBLP:conf/kdd/ChenL00WYC22} incorporate visual features to guide relational feature learning while weighting valuable attributes for alignment. 
Lin et al. \cite{DBLP:conf/coling/LinZWSW022} further improve intra-modal learning with contrastive learning. 
Shi et al. \cite{DBLP:journals/dase/WangSYZLZ23} filter out mismatched images with pre-defined ontologies and an image type classifier.
Chen et al. \cite{chen2023meaformer} dynamically predict the mutual modality weights for entity-level modality fusion and alignment.

These approaches substantiate that visual information indeed contributes positively to EA.
However, we notice that all of them are based on two ideal assumptions: {{(\rmnum{1})}} Entities and images have a one-to-one correspondence, meaning that a single image sufficiently encapsulates and conveys all the information about an entity. {{(\rmnum{2})}} Images are always available, implying that an entity consistently possesses a corresponding image.

\begin{figure*}[!htbp]
  \centering
  \includegraphics[width=1.0\linewidth]{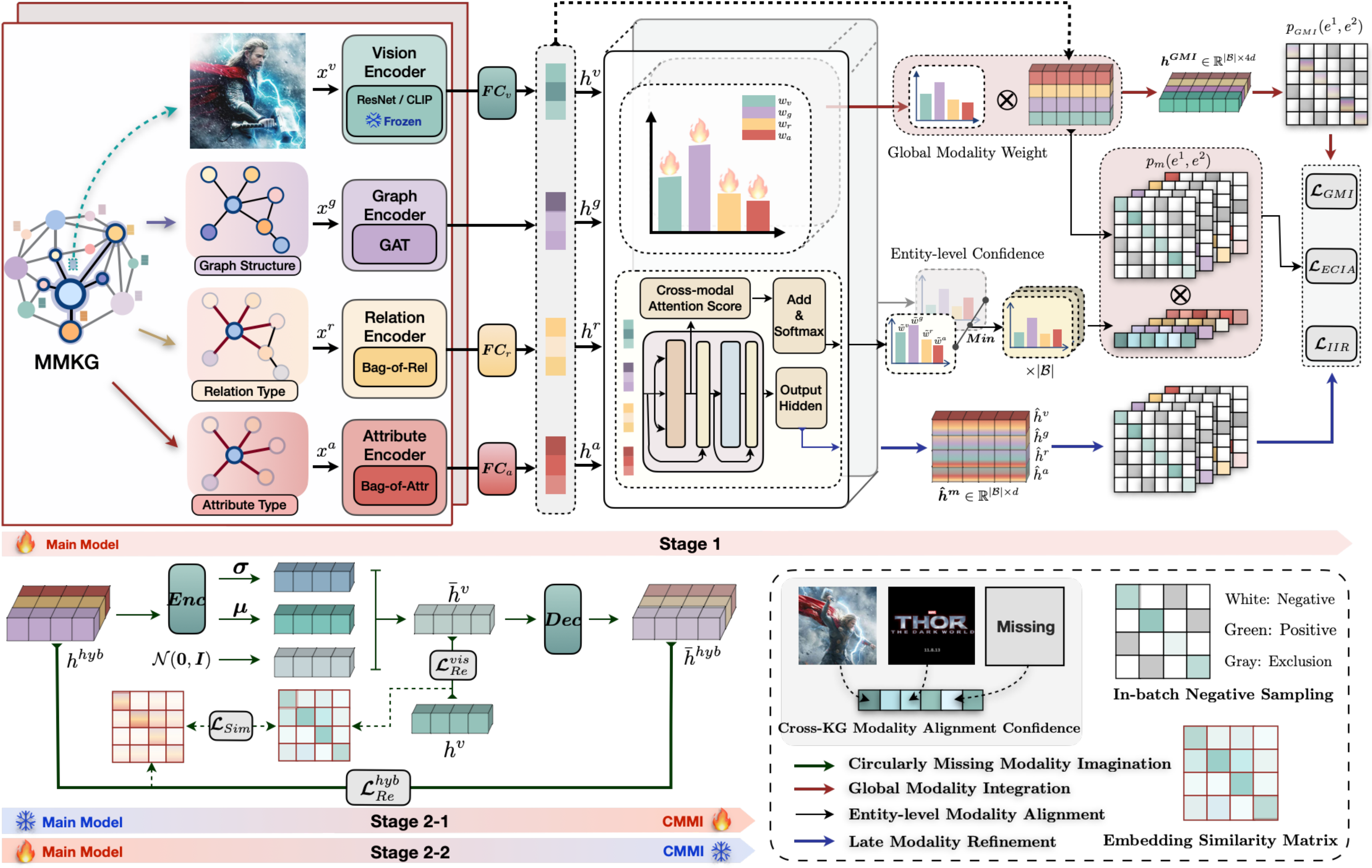}
  \caption{The overall framework of UMAEA. }
  \label{fig:model}
\end{figure*}

In real-world KGs, the noise is an inherent issue. Even for the standard MMEA datasets \cite{DBLP:conf/semweb/SunHL17,DBLP:conf/esws/LiuLGNOR19,DBLP:conf/ksem/ChenLWXWC20,DBLP:conf/aaai/0001CRC21}, they are hard to satisfy those two ideal conditions mentioned above. 
Consequently, we focus on two more pragmatic and demanding issues: {{(\rmnum{1})}} {In MMKGs, entity images might be missing uncertainly, implying a varying degree of image absence.}  {{(\rmnum{2})}} {In MMKGs, images of the entities could be uncertainly ambiguous, suggesting that a single entity might have heterogeneous visual representations. }
To tackle these challenges, we present a benchmark consisting of seven datasets on which extensive experiments are conducted, and introduce our model UMAEA against these problems.

\subsubsection{Incomplete Multi-modal Learning}
aims to tackle classification or reconstruction tasks, like multi-modal emotion recognition \cite{DBLP:conf/acl/ZhaoLJ20} and cross-modal retrieval \cite{DBLP:conf/mm/JingLZLYH20},  by leveraging information from available modalities when one modality is missing (e.g,  a tweet may only have images or text content).
In multi-modal alignment tasks, missing modality significantly impacts the performance as the symmetry of paired multi-modal data leads to noise accumulation when it is uncertain which side has modality incompleteness, 
further hindering model training. 
Prior MMEA studies \cite{DBLP:conf/aaai/0001CRC21,DBLP:conf/kdd/ChenL00WYC22,DBLP:conf/coling/LinZWSW022,chen2023meaformer} calculate mean and variance from available visual features, enabling random generation of those incomplete features using a normal distribution. In this paper, we develop an adaptive method for optimal training under the conditions with uncertainly missing or noisy visual modality, meanwhile providing a comprehensive benchmark.

\section{Method}
\subsection{Preliminaries}
We define a MMKG as a five-tuple $\mathcal{G}$$=$$\{\mathcal{E}, \mathcal{R}, \mathcal{A}, \mathcal{V}, \mathcal{T}\}$, where
$\mathcal{E}, \mathcal{R}, \mathcal{A}$ and $\mathcal{V}$ denote the sets of entities, relations, attributes, and images, respectively. $\mathcal{T}$ $\subseteq$ $\mathcal{E} \times \mathcal{R} \times\mathcal{E}$ is the set of relation triples.
Given two MMKGs $\mathcal{G}_1$ $=$ $\{\mathcal{E}_1, \mathcal{R}_1, \mathcal{A}_1, \mathcal{V}_1, \mathcal{T}_1\}$ and $\mathcal{G}_2$ $=$ $\{\mathcal{E}_2, \mathcal{R}_2, \mathcal{A}_2, \mathcal{V}_2, \mathcal{T}_2\}$, MMEA aims to discern each entity pair ($e^1_i$, $e^2_i$), $e^1_i \in \mathcal{E}_1$, $e^2_i \in \mathcal{E}_2$ where $e^1_i$ and $e^2_i$ correspond to an identical real-world entity $e_i$.
For clarity, we omit the superscript symbol denoting the source KG of an entity in our context, except when explicitly required in statements or formulas.
A set of pre-aligned entity pairs is provided, which is proportionally divided into a training set (i.e., seed alignments $\mathcal{S}$) and a testing set $\mathcal{S}_{te}$ based on a given seed alignment ratio ($R_{sa}$). 
We denote $\mathcal{M}=\{g, r, a, v\}$ as the set of available modalities.
Commonly, 
in typical KG datasets for MMEA, each entity is associated with multiple attributes and  $0$ or $1$ image, and the proportion ($R_{img}$) of entities containing images is uncertain (e.g., 67.58$\%$ in DBP15K$_{JA\text{-}EN}$ \cite{DBLP:conf/aaai/0001CRC21}). 
In this study, in order to facilitate a comprehensive evaluation, dataset MMEA-UMAM is proposed where we  define $R_{img}$ as a controlled variable for  benchmarking.

\subsection{Multi-modal Knowledge Embedding} 
\subsubsection{Graph Structure Embedding.} 
Let $x_i^g \in \mathbb{R}^d$ represent the  randomly initialized graph embedding of entity $e_i$ where $d$ is the predetermined hidden dimension. 
We employ the Graph Attention Network (GAT) \cite{DBLP:conf/iclr/VelickovicCCRLB18} with two attention heads and two layers to capture the structural information of $\mathcal{G}$, equipped with a diagonal weight matrix \cite{DBLP:journals/corr/YangYHGD14a} $\bm{W}_g \in \mathbb{R}^{d \times d}$ for linear transformation. We define
 $h_i^g = GAT(\bm{W}_g, \bm{M}_g; x_i^g)\,$,
where $\bm{M}_g$ denotes to the graph adjacency matrix. 

\subsubsection{Relation, Attribute, and Visual Embedding.} 
To mitigate the information contamination arising from blending relation / attribute representations in GNN-like networks \cite{DBLP:conf/aaai/0001CRC21}, we employ separate fully connected layers, parameterized by $\bm{W}_m \in \mathbb{R}^{d_m \times d}$, 
for embedding space harmonization via
 $h_i^m$ $=$ $FC_m(\bm{W}_m, x_i^m)\,$,
where $m \in \{r, a, v\}$ and $r$, $a$, $v$, represent relation, attribute, visual modalities, respectively. 
Furthermore,  $x_i^m \in \mathbb{R}^{d_m}$ denotes the input feature of entity $e_i$ for the corresponding modality $m$.
We follow Yang et al. \cite{DBLP:conf/emnlp/YangZSLLS19} to use the bag-of-words features for relation ($x^r$) and attribute ($x^a$) representations (see Section \ref{sec:detail} for details).
While for the visual modality, we employ a pre-trained (frozen) visual model as the encoder ($Enc_v$) to obtain the visual embeddings $x^v_i$ for each available image of the entity $e_i$.
For entities without image data, we generate random image features using a normal distribution parameterised by the mean and standard deviation of other available images \cite{DBLP:conf/aaai/0001CRC21,DBLP:conf/kdd/ChenL00WYC22,DBLP:conf/coling/LinZWSW022,chen2023meaformer}.  

\subsection{Multi-scale Modality Hybrid}
This section describes the detailed architecture of the multi-scale modality hybrid for aligning multi-modal entities between MMKGs. The model comprises three modality alignment modules operating at different scales, each associated with a training objective as depicted in Figure \ref{fig:model}.

\subsubsection{Global Modality Integration} (GMI) emphasizes global alignment for each multi-modal entity pair, where the multi-modal embeddings for an entity are first concatenated and then aligned using a learnable global weight, allowing the model to adaptively learn the relative quality of each modality across two MMKGs. 
Let $w_m$ be the global weight for modality $m$. We formulate the GMI joint embedding ${h}^{GMI}_i$ for entity $e_i$ as: 
\begin{align} \label{eq:cat}
    {h}^{GMI}_i = \bigoplus\nolimits_{m \in \mathcal{M}}[w_m{h}_i^m]\, ,
\end{align} 
where $\bigoplus$ refers to the vector concatenation operation.
To enhance model's sensitivity to feature differences between unaligned entities, we introduce a unified entity alignment contrastive learning framework, inspired by Lin et al. \cite{DBLP:conf/coling/LinZWSW022}, to consolidate the training objectives of the modules. For each entity pair ($e_i^1$,$e_i^2$) in $\mathcal{S}$, we define $\mathcal{N}^{ng}_i$ $=$ $\{e^1_j|\forall e^1_j \in \mathcal{E}_1, j \neq i\}$ $\cup$ $\{e^2_j|\forall e^2_j \in \mathcal{E}_2, j \neq i\}$ as its negative entity set.
To improve efficiency, we adopt the in-batch negative sampling strategy \cite{DBLP:conf/icml/ChenK0H20}, restricting the sampling scope of  $\mathcal{N}^{ng}_i$ to the mini-batch $\mathcal{B}$.
Concretely,
we define the alignment probability distribution as follows:
\begin{equation}\label{eq:macl}
    p_m(e^1_i, e^2_i) =  \frac{\gamma_m(e^1_i, e^2_i)}{\gamma_m(e^1_i, e^2_i) + \sum\nolimits_{e_j \in \mathcal{N}^{ng}_i}\gamma_m(e^1_i, e_j)} \, ,
\end{equation}
where $\gamma_m(e_i, e_j)$ $=$ $\exp({h^{m\top}_{i}}{h^m_{j}}/\tau)$ and $\tau$ represents the temperature hyper-parameter.
To account for the alignment direction of entity pairs in \eqref{eq:macl}, we establish a bi-directional alignment objective as:
\begin{equation}\label{eq:clloss}
    \mathcal{L}_m = - \mathbb{E}_{i \in \mathcal{B}}\, \log[\,p_m(e^1_i, e^2_i)+p_m(e^2_i, e^1_i)\,]/2 \,,
\end{equation}
where $m$ denotes a modality or an embedding type. We denote the training objective as $\mathcal{L}_{GMI}$ when the GMI join embedding is used, i.e., $\gamma_{GMI}(e_i, e_j)$ is set to $\exp({h^{{GMI}\top}_{i}}{h^{GMI}_{j}}/\tau)$.

We note that the global adaptive weighting allows the model to capitalize on high-quality modalities while minimizing the impact of low-quality modalities, such as the redundant information within attributes / relations, and noise within images. Concurrently, it ensures the preservation of valuable information to a certain extent, ultimately contributing to the stability of the alignment process.

\subsubsection{Entity-level Modality Alignment} aims to perform instance-level modality weighting and alignment, utilizing minimum cross-KG  confidence measures from seed alignments to constrain the modality alignment objectives.
It allows the model to dynamically assign lower training weights to missing or ambiguous modality information, thereby reducing the risk of encoder misdirection arising from uncertainties. 
To achieve this, we follow Chen et al. \cite{chen2023meaformer} to adapt the vanilla Transformer \cite{DBLP:conf/nips/VaswaniSPUJGKP17} for two types of sub-layers:  
the multi-head cross-modal attention (MHCA) block and the fully connected feed-forward networks (FFN).

Specifically, MHCA operates its attention function across $N_h$ parallel heads. The $i$-th head is parameterized by modally shared matrices $\bm{W}_q^{(i)}$, $\bm{W}_k^{(i)}$, $\bm{W}_v^{(i)}$ $\in \mathbb{R}^{d \times d_h}$, transforming the multi-modal input $h^m$ into modal-aware query ${Q}^{(i)}_m$, key ${K}^{(i)}_m$, and value ${V}^{(i)}_m$ in $\mathbb{R}^{d_h}$ ($d_h=d/N_h$):
\begin{equation}
{Q}^{(i)}_m, {K}^{(i)}_m, {V}^{(i)}_m =h^m \bm{W}_q^{(i)}, h^m \bm{W}_k^{(i)}, h^m \bm{W}_v^{(i)} \,.\\
\end{equation}
MHCA generates the following output for a given feature of modality $m$:
\begin{align}
       \operatorname{MHCA}(h^m) & =\bigoplus\nolimits_{i=1}^{N_h}\operatorname{head}_i^m \cdot \bm{W}{o} \,, \\
		\operatorname{head}_{\mathrm{i}}^m & = \sum\nolimits_{j \in \mathcal{M}} \beta^{(i)}_{mj}{V}^{(i)}_j \,,
\end{align}
where $\bm{W}_{o}$ $\in \mathbb{R}^{d \times d}$. The attention weight ($\beta_{m j}$)
 between an entity's modality $m$ and $j$  in each head  is calculated as:
\begin{equation}
    \beta_{m j}=\frac{\exp (Q_m^\top K_j / \sqrt{d_h} )}{\sum_{i \in \mathcal{M}} \exp (Q_m^\top K_i / \sqrt{d_h})}\,.
\end{equation}
Besides, layer normalization (LN) and residual connection (RC) are incorporated to stabilize training:
\begin{equation}
\hat{h}^m = LayerNorm(\operatorname{MHCA}(h^m) + h^m) \,.
\end{equation}

The FFN consists of two linear transformation layers and a ReLU activation function with LN and RC applied afterwards:
\begin{align}
\operatorname{FFN}(\hat{h}^m) & = ReLU(\hat{h}^m\bm{W}_{1} + b_{1})\bm{W}_{2} +b_{2} \,, \\
\label{eq:hidden}
\hat{h}^m & \gets LayerNorm(\operatorname{FFN}(\hat{h}^m) + \hat{h}^m) \,,
\end{align}
where $\bm{W}_{1}$ $\in \mathbb{R}^{d \times d_{in}}$ and $\bm{W}_{2}$ $\in \mathbb{R}^{d_{in} \times d}$.
Notably, 
we define the entity-level confidence $\tilde{w}^m$ for each modality $m$ as:
\begin{equation}
    \tilde{w}^m = \frac{\exp(\sum\nolimits_{j \in \mathcal{M}} \sum\nolimits_{i=0}^{N_h}  \beta^{(i)}_{mj}/\sqrt{|\mathcal{M}| \times N_h})}{\sum\nolimits_{k \in \mathcal{M}}\exp(\sum\nolimits_{j \in \mathcal{M}} \sum\nolimits_{i=0}^{N_h}  \beta^{(i)}_{kj}\sqrt{|\mathcal{M}| \times N_h})}\,,
\end{equation}
which captures crucial inter-modal interface information and adaptively adjusts model's cross-KG alignment confidence for different modalities from each entity.
To facilitate learning these dynamic confidences and incorporating them into the training process, we devise two distinct training objectives: $\mathcal{L}_{ECIA}$ and $\mathcal{L}_{IIR}$.
The first objective is \emph{explicit confidence-augmented intra-modal alignment} (ECIA), while the second is \emph{implicit inter-modal refinement} (IIR), which will be discussed in the following subsection.
For the ECIA, we design the following training target which is the variation of 
 Equation \eqref{eq:clloss}:
\begin{align}
    \mathcal{L}_{ECIA} & = \sum\nolimits_{m \in \mathcal{M}}\widetilde{\mathcal{L}}_m\,\, , \\
    \widetilde{\mathcal{L}}_m = - \mathbb{E}_{i \in \mathcal{B}}\,\log[\,\phi_m(e^1_i, e^2_i& )*(p_m(e^1_i, e^2_i)+p_m(e^2_i, e^1_i))\,]/2 \,.
\end{align}
Considering the symmetric nature of EA and the varying quality of aligned entities and their modality features within each KG, we employ the minimum confidence value to minimize errors. For example, $e^1_i$ may possess high-quality image data while $e^2_i$ lacks image information, as illustrated in Figure \ref{fig:Introcase}. In such cases, using the original objective for feature alignment will inadvertently align meaningful features with random noise, thereby disrupting the encoder training process.
To mitigate this issue, we define $\phi_m(e^1_i, e^2_i)$ as the minimum confidence value for entities $e^1_i$ and $e^2_i$ in modality $m$, calculated by
	$\phi_m(e_i, e_j) = Min(\tilde{w}_i^m, \tilde{w}_j^m)\,$.

\subsubsection{Late Modality Refinement} \label{sec:iir}
 leverages the transformer layer outputs to 
 further enhance the entity-level adaptive modality alignment through an \emph{implicit inter-modal refinement} (IIR) objective, enabling the refinement of attention scores by directly aligning the output hidden states.
Concretely, we define the hidden state embedding of modality $m$ for entity $e_i$ as $\hat{h}^m$, following  Equation \eqref{eq:hidden}. We define:
\begin{equation}
 \mathcal{L}_{IIR} = \sum\nolimits_{m \in \mathcal{M}}\widehat{\mathcal{L}}_m\,,
\end{equation}
where $\widehat{\mathcal{L}}_m$ is also a variant of $\mathcal{L}_m$,  as illustrated in Equation \eqref{eq:clloss}, with only the following modification: 
$\widehat{\gamma}_m(e_i, e_j) = \exp(\hat{h}^{m\top}_{i}{\hat{h}^m_{j}}/\tau)$.

As depicted in Figure \ref{fig:model}, we designate the entire process so far as the first stage of our (main) model, with the training objective formulated as:
\begin{equation}
\mathcal{L}_{1} = \mathcal{L}_{GMI} + \mathcal{L}_{ECIA} + \mathcal{L}_{IIR} \,.
\end{equation}


\subsection{Circularly Missing Modality Imagination.}
Note that our primary target of the first stage is to alleviate the impact of modality noise and incompleteness on the alignment process throughout training. Conversely,  the second stage draws inspiration from VAE \cite{DBLP:journals/corr/KingmaW13,DBLP:conf/nips/SohnLY15} and CycleGAN \cite{DBLP:conf/iccv/ZhuPIE17}, which accentuates generative modeling and unsupervised domain translation. Expanding upon these ideas, we develop our circularly missing modality imagination (CMMI) module, aiming to enable the model to proactively complete missing modality information. 

To reach our goal, we develop a variational multi-modal autoencoder framework, allowing the hidden layer output between the encoder $MLP_{Enc}$ and decoder $MLP_{Dec}$ (parameterized by $\bm{W}_{Enc} \in \mathbb{R}^{3d \times 2d}$ and $\bm{W}_{Dec} \in \mathbb{R}^{d \times 3d}$, respectively) to act as an imagined pseudo-visual feature $\bar{h}^{v}_i$, using reparameterization strategy \cite{DBLP:journals/corr/KingmaW13} with tri-modal hybrid feature ${h}^{hyb}_i = [{h}_i^r \oplus {h}_i^a \oplus {h}_i^g]$ as the input:
\begin{align}
\,\,\,\,\,\, [\,{\mu}_i \oplus \log(\sigma_i)^2\,] & = MLP_{Enc}({h}^{hyb}_i) \,, \\
\,\,\,\,\,\, \bar{h}^{v}_i & = z  \odot {\sigma}_i + {\mu}_i \,,\,\,\, z \sim \mathcal{N}(\bm{0},\bm{I}) \,, \\
\,\,\,\,\,\, \bar{h}^{hyb}_i & = MLP_{Dec}(\bar{h}^{v}_i) \, .
\end{align}
Concretely, two reconstruction objectives $\mathcal{L}_{Re}^{vis}$ and $\mathcal{L}_{Re}^{hyb}$ are utilized to minimize $|{h}^{hyb}_i - \bar{h}^{hyb}_i|$ and $|h^{v}_i - \bar{h}^{v}_i|$, where $h^{v}_i$ represents the real image feature.
Besides, we adhere to the standard VAE algorithm \cite{DBLP:journals/corr/KingmaW13} to regularize the latent space by encouraging it to be similar to a Gaussian distribution through minimizing the Kullback–Leibler (KL) divergence:
\begin{equation}
\mathcal{L}_{KL} = \mathbb{E}_{i \in \mathcal{\bar{B}}}\, ((\mu_i)^2 + (\sigma_i)^2 - \log(\sigma_i)^2 - 1)/2 \,,
\end{equation}
where $\mathcal{\bar{B}}$ refers to those entities with complete images within a mini-batch.

Furthermore, we exploit the internal embedding similarity matrix obtained from the  hybrid embeddings ${h}^{hyb}$, and distill this information into the virtual image feature similarity matrix based on $\bar{h}^{v}$:
\begin{equation}
\mathcal{L}_{Sim} = \mathbb{E}_{i \in \mathcal{\bar{B}}}\, D_{KL}(p_{hyb}(e^1_i, e^2_i)||\bar{p}_{v}(e^1_i, e^2_i))\,,
\end{equation}
where $p_{hyb}$ and $\bar{p}_{v}$ all follow Equation \eqref{eq:macl} with $\gamma_{hyb}(e_i, e_j)$ $=$ $\exp({h^{hyb\top}_{i}}{h^{hyb}_{j}}/\tau)$ and $\bar{\gamma}_v(e_i, e_j)$ $=$ $\exp({\bar{h}^{v\top}_{i}}{\bar{h}^v_{j}}/\tau)$. 
This strategy not only curbs the overfitting of visible visual modalities in the autoencoding process, but also emphasizes the differences between distinct characteristics. Crucially, the knowledge mapping of original tri-modal hybrid features to the visual space  is maximally preserved, thereby mitigating modal collapse when most of the visual content is missing and the noise is involved. The final loss in stage two is formulated as:
\begin{equation}
\mathcal{L}_{2} = \mathcal{L}_{KL} + \mathcal{L}_{Re}^{vis} + \mathcal{L}_{Re}^{hyb} + \mathcal{L}_{Sim} \,.
\end{equation}

\subsection{Training Details}
\subsubsection{Pipeline.}
As previously mentioned, the training process consists of two stages. In the first stage, the primary model components are trained independently, while in the second stage, the CMMI module is additionally incorporated. The training objective $\mathcal{L}$ is defined as follows: 
\begin{align}
Stage~1: \mathcal{L} & \gets \mathcal{L}_1 \,, \\
Stage~2\text{-}1/2\text{-}2: \mathcal{L} & \gets \mathcal{L}_1 + \mathcal{L}_2 \,,
\end{align}
where the second stage is further divided into two sub-stages.
Concretely, in order to stabilize model training and avoid knowledge forgetting  caused by the cold-start of module insertion  \cite{DBLP:journals/corr/abs-2304-14178}, as shown in Figure \ref{fig:model}, the models from stage 1 (i.e., main model) are frozen to facilitate CMMI training when entering stage 2-1. While in stage 2-2, the CMMI is frozen and the main model undergoes further refinement to establish the entire pipeline.
This process is easy to implement, just by switching the range of learnable parameters during model training.

\subsubsection{Entity Representation.}
During evaluation, we replace the original random vectors with the generated $\mu_i$ for those entities without images. While in the second training stage, we employ the pseudo-visual embedding $\bar{h}^v_i$ (rather than $\mu_i$) as a substitute as 
 we observe that actively introducing noise during training could introduce randomness and uncertainty into the reconstruction process, which has been demonstrated to be beneficial in learning sophisticated distributions and enhances the model's robustness \cite{DBLP:conf/iclr/LeeNYH20}.
Furthermore,  we select ${h}^{GMI}_i$, as formulated in Equation \eqref{eq:cat}, for the final multi-modal entity representation. 


\section{Experiment}
\subsection{Experiment Setup}
To guarantee a fair assessment, we use a total of seven MMEA datasets derived from three major categories ({bilingual}, {monolingual}, and {high-degree}), with two representative  pre-trained visual encoders (ResNet-152 \cite{DBLP:conf/cvpr/HeZRS16} and CLIP \cite{DBLP:conf/icml/RadfordKHRGASAM21}), and evaluated the performance of {four} models under two distinct settings ({standard} (non-iterative) and {iterative}). In this research, we intentionally set aside the surface modality (literal information) to focus on understanding the effects of absent visual modality on model performance.

\subsubsection{Datasets.}
DBP15K \cite{DBLP:conf/semweb/SunHL17} contains three  datasets ($R_{sa}=0.3$) built from the multilingual versions of DBpedia, including DBP15K$_{ZH\text{-}EN}$, DBP15K$_{JA\text{-}EN}$ and DBP15K$_{FR\text{-}EN}$. 
We adopt their multi-model variants \cite{DBLP:conf/aaai/0001CRC21} with entity-matched images attached.
Besides, four Multi-OpenEA datasets ($R_{sa}=0.2$) \cite{DBLP:journals/corr/abs-2302-08774} are used, which are the multi-modal variants of the OpenEA benchmarks \cite{DBLP:journals/pvldb/SunZHWCAL20} with entity images achieved by searching the entity names through the Google search engine. We include two bilingual datasets \{ EN-FR-15K, EN-DE-15K \} and two monolingual datasets \{ D-W-15K-V1, D-W-15K-V2 \}, where V1 and V2 denote two versions with distinct average relation degrees.
To create our \textbf{MMEA-UMVM} (uncertainly missing visual modality) datasets, we perform random image dropping on MMEA datasets. Specifically, we randomly discard entity images to achieve varying degrees of visual modality missing, ranging from 0.05 to the maximum $R_{img}$ of the raw datasets with a step of 0.05 or 0.1. 
Finally, we get a total number of 97 data split. 
See appendix \footnote{The appendix is attached with the arXiv version of this paper.} for more details. 


\subsubsection{Iterative Training.}
Following Lin et al. \cite{DBLP:conf/coling/LinZWSW022}, we adopt a probation technique for iterative training. The probation can be viewed as a buffering mechanism, which maintains a temporary cache to store cross-graph mutual nearest entity pairs from the testing set.
Concretely, every $K_e$ (where $K_e = 5$) epochs, we propose cross-KG entity pairs that are mutual nearest neighbors in the vector space and add them to a candidate list $\mathcal{N}^{cd}$. 
Furthermore, an entity pair in $\mathcal{N}^{cd}$ will be added into the training set if it remains a mutual nearest neighbour for $K_s$ ($=$ $10$) consecutive rounds.

\subsubsection{Baselines.}
Six prominent EA algorithms proposed in recent years are selected as our baseline comparisons, excluding the surface information for a parallel evaluation.
We further collect 3 latest MMEA methods as the strong baselines, including EVA \cite{DBLP:conf/aaai/0001CRC21}, MSNEA \cite{DBLP:conf/kdd/ChenL00WYC22}, and MCLEA \cite{DBLP:conf/coling/LinZWSW022}. Particularly, we reproduce them with their original pipelines unchanged in our benchmark.
\begin{table}[!htbp]
    \centering
	\tabcolsep=0.3cm
    \renewcommand\arraystretch{1.0}
    \caption{{Non-iterative} results of four models with ``w/o CMMI'' setting indicating the absence of the stage-2. 
    The best results within the baselines are marked with \underline{underline}, and we highlight our results with \textbf{bold} when we achieve SOTA.} 
    \resizebox{0.98\linewidth}{!}{
    \begin{tabular}{@{}l|l|ccc|ccc|ccc|ccc}
        \toprule
        & \multirow{2}*{\makebox[1.8cm][c]{Models}} & \multicolumn{3}{c|}{$R_{img}$ $=$ $0.05$} & \multicolumn{3}{c|}{$R_{img}$ $=$ $0.2$} & \multicolumn{3}{c|}{$R_{img}$ $=$ $0.4$} & \multicolumn{3}{c}{$R_{img}$ $=$ $0.6$} \\
        & & {\scriptsize H@1} & {\scriptsize H@10} & {\scriptsize MRR} & {\scriptsize H@1} & {\scriptsize H@10} & {\scriptsize MRR} & {\scriptsize H@1} & {\scriptsize H@10} & {\scriptsize MRR} & {\scriptsize H@1} & {\scriptsize H@10} & {\scriptsize MRR} \\
        \midrule
        \parbox[t]{2mm}{\multirow{6}{*}{\rotatebox[origin=c]{90}{DBP15K$_{ZH-EN}$}}} 
        & MSNEA {\footnotesize {\cite{DBLP:conf/kdd/ChenL00WYC22}}} & .413 & .722 & .517 & .411 & .725 & .518 & .446 & .743 & .546 & .520 & .786 & .611  \\
        & EVA {\footnotesize \cite{DBLP:conf/aaai/0001CRC21}} &
        {.623} & {.878} & {.715} & \underline{.624} & \underline{.878} & \underline{.716} & \underline{.623} & \underline{.875} & \underline{.714} & .625 & .876 & .717 \\
        & MCLEA {\footnotesize {\cite{DBLP:conf/coling/LinZWSW022}}}  &
        \underline{.638} & \underline{.905} & \underline{.732} & {.588} & {.865} & {.686} & {.611} & {.874} & {.704} & \underline{.661} & \underline{.896} & \underline{.744} \\
        & \CC\textbf{w/o CMMI}  
        & \CC{.703} & \CC{.934} & \CC{.787} & \CC{.710} & \CC{.937} & \CC{.793} & \CC{.721} & \CC{.939} & \CC{.801} & \CC{.753} & \CC{.949} & \CC{.825}  \\
        & \CC\textbf{UMAEA}  &
        \CC\textbf{.720} & \CC\textbf{.938} & \CC\textbf{.800} & \CC\textbf{.727} & \CC\textbf{.941} & \CC\textbf{.806} & \CC\textbf{.727} & \CC\textbf{.941} & \CC\textbf{.806} & \CC\textbf{.758} & \CC\textbf{.951} & \CC\textbf{.829} \\
        \cmidrule(lr){2-14}
        & {Improve {$\uparrow$}}  
        & \small{ 8.2$\%$} & \small{ 3.3$\%$} & \small{ .068} & \small{ 10.3$\%$} & \small{ 6.3$\%$} & \small{ .090} & \small{ 10.4$\%$} & \small{ 6.6$\%$} & \small{ .092} & \small{ 9.7$\%$} & \small{ 5.5$\%$} & \small{ .085}  \\
        \midrule
        \parbox[t]{2mm}{\multirow{6}{*}{\rotatebox[origin=c]{90}{DBP15K$_{JA-EN}$}}}
        & MSNEA {\footnotesize {\cite{DBLP:conf/kdd/ChenL00WYC22}}}  & .313 & .643 & .425 & .311 & .644 & .422 & .369 & .678 & .472 & .480 & .744 & .569 \\
        & EVA {\footnotesize \cite{DBLP:conf/aaai/0001CRC21}}  & \underline{.615} & .877 & \underline{.708} & \underline{.616} & \underline{.877} & \underline{.710} & \underline{.616} & \underline{.878} & \underline{.711} & .624 & .881 & .716 \\
        & MCLEA {\footnotesize {\cite{DBLP:conf/coling/LinZWSW022}}}  &
        {.599} & \underline{.897} & {.706} & {.579} & {.846} & {.675} & {.613} & {.867} & {.703} & \underline{.686} & \underline{.898} & \underline{.761} \\
        & \CC\textbf{w/o CMMI}  
        & \CC{.708} & \CC{.943} & \CC{.794} & \CC{.712} & \CC{.947} & \CC{.798} & \CC{.730} & \CC{.950} & \CC{.810} & \CC{.772} & \CC{.962} & \CC{.843}  \\
        & \CC\textbf{UMAEA}   &
        \CC\textbf{.725} & \CC\textbf{.949} & \CC\textbf{.807} & \CC\textbf{.726} & \CC\textbf{.949} & \CC\textbf{.808} & \CC\textbf{.732} & \CC\textbf{.952} & \CC\textbf{.813} & \CC\textbf{.775} & \CC\textbf{.963} & \CC\textbf{.845} \\
        \cmidrule(lr){2-14}
        & {Improve {$\uparrow$}}  
        & \small{ 11.0$\%$} & \small{ 5.2$\%$} & \small{ .099} & \small{ 11.0$\%$} & \small{ 7.2$\%$} & \small{ .098} & \small{ 11.6$\%$} & \small{ 7.4$\%$} & \small{ .102} & \small{ 8.9$\%$} & \small{ 6.5$\%$} & \small{ .084}  \\
		\midrule
		\parbox[t]{2mm}{\multirow{6}{*}{\rotatebox[origin=c]{90}{DBP15K$_{FR-EN}$}}} 
		& MSNEA {\footnotesize {\cite{DBLP:conf/kdd/ChenL00WYC22}}}  & .297 & .690 & .427 & .304 & .690 & .428 & .360 & .710 & .474 & .478 & .772 & .574 \\
		& EVA {\footnotesize \cite{DBLP:conf/aaai/0001CRC21}}  & .624 & .895 & .720 & \underline{.624} & \underline{.895} & \underline{.720} & \underline{.626} & \underline{.898} & \underline{.721} & .634 & .900 & .728 \\
        & MCLEA {\footnotesize {\cite{DBLP:conf/coling/LinZWSW022}}}  &
        \underline{.634} & \underline{.930} & \underline{.741} & {.582} & {.863} & {.682} & {.601} & {.879} & {.702} & \underline{.675} & \underline{.901} & \underline{.757} \\
        & \CC\textbf{w/o CMMI}  
        & \CC{.727} & \CC{.956} & \CC{.813}& \CC{.733} & \CC{.960} & \CC{.817} & \CC{.746} & \CC{.961} & \CC{.828} & \CC{.790} & \CC{.968} & \CC{.857}\\
        & \CC\textbf{UMAEA}   &
        \CC\textbf{.752} & \CC\textbf{.970} & \CC\textbf{.830} & \CC\textbf{.755} & \CC\textbf{.960} & \CC\textbf{.832} & \CC\textbf{.763} & \CC\textbf{.962} & \CC\textbf{.838} & \CC\textbf{.792} & \CC\textbf{.970} & \CC\textbf{.859} \\
        \cmidrule(lr){2-14}
        & {Improve {$\uparrow$}}  
        & \small{ 11.8$\%$} & \small{ 4.0$\%$} & \small{ .089} & \small{ 13.1$\%$} & \small{ 6.7$\%$} & \small{ .112} & \small{ 13.7$\%$} & \small{ 6.4$\%$} & \small{ .117} & \small{ 11.7$\%$} & \small{ 6.9$\%$} & \small{ .102}  \\
		\midrule
		\parbox[t]{2mm}{\multirow{6}{*}{\rotatebox[origin=c]{90}{OpenEA$_{EN-FR}$}}} 
		& MSNEA {\footnotesize {\cite{DBLP:conf/kdd/ChenL00WYC22}}}  
		& .200 & .431 & .278 & .213 & .439 & .290 & .260 & .477 & .334 & .360 & .560 & .427 \\
		& EVA {\footnotesize \cite{DBLP:conf/aaai/0001CRC21}}  
		& .528 & .833 & .634 & {.533} & {.835} & {.638} & \underline{.539} & {.835} & \underline{.642} & .547 & .830 & .647 \\
        & MCLEA {\footnotesize {\cite{DBLP:conf/coling/LinZWSW022}}}  
        & \underline{.545} & \underline{.852} & \underline{.653} & \underline{.547} & \underline{.852} & \underline{.655} & {.531} & \underline{.839} & {.637} & \underline{.597} & \underline{.852} & \underline{.688} \\
        & \CC\textbf{w/o CMMI}  
        & \CC{.587} & \CC{.893} & \CC{.695} & \CC{.590} & \CC{.893} & \CC{.697} & \CC{.614} & \CC\textbf{.900} & \CC{.715} & \CC{.664} & \CC{.912} & \CC{.753}  \\
        & \CC\textbf{UMAEA}   &
        \CC\textbf{.605} & \CC\textbf{.898} & \CC\textbf{.708} & \CC\textbf{.604} & \CC\textbf{.896} & \CC\textbf{.708} & \CC\textbf{.618} & \CC{.899} & \CC\textbf{.718} & \CC\textbf{.665} & \CC\textbf{.914} & \CC\textbf{.753} \\
        \cmidrule(lr){2-14}
        & {Improve {$\uparrow$}}  
        & \small{ 6.0$\%$} & \small{ 4.6$\%$} & \small{ .055} & \small{ 5.7$\%$} & \small{ 4.4$\%$} & \small{ .053} & \small{ 7.9$\%$} & \small{ 6.1$\%$} & \small{ .076} & \small{ 6.8$\%$} & \small{ 6.2$\%$} & \small{ .065}  \\
        \midrule
		\parbox[t]{2mm}{\multirow{6}{*}{\rotatebox[origin=c]{90}{OpenEA$_{EN-DE}$}}} 
		& MSNEA {\footnotesize {\cite{DBLP:conf/kdd/ChenL00WYC22}}}  & .242 & .486 & .323 & .253 & .495 & .333 & .309 & .542 & .387 & .412 & .622 & .484 \\
		& EVA {\footnotesize \cite{DBLP:conf/aaai/0001CRC21}}  & .717 & .917 & .787 & {.718} & \underline{.918} & {.788} & \underline{.721} & \underline{.920} & \underline{.791} & .734 & \underline{.921} & .800 \\
        & MCLEA {\footnotesize {\cite{DBLP:conf/coling/LinZWSW022}}}  
        & \underline{.723} & \underline{.918} & \underline{.791} & \underline{.721} & {.915} & \underline{.789} & {.697} & {.907} & {.771} & \underline{.745} & {.906} & \underline{.803} \\
        & \CC\textbf{w/o CMMI}  
        & \CC{.752} & \CC{.938} & \CC{.818} & \CC{.757} & \CC{.941} & \CC{.822} & \CC{.771} & \CC{.946} & \CC{.833} & \CC\textbf{.804} & \CC{.954} & \CC{.858}  \\
        & \CC\textbf{UMAEA} &
        \CC\textbf{.757} & \CC\textbf{.942} & \CC\textbf{.823} & \CC\textbf{.759} & \CC\textbf{.943} & \CC\textbf{.824} & \CC\textbf{.774} & \CC\textbf{.947} & \CC\textbf{.835} & \CC\textbf{.804} & \CC\textbf{.957} & \CC\textbf{.860} \\
        \cmidrule(lr){2-14}
        & {Improve {$\uparrow$}}  
        & \small{ 3.4$\%$} & \small{ 2.4$\%$} & \small{ .032} & \small{ 3.8$\%$} & \small{ 2.5$\%$} & \small{ .035} & \small{ 5.3$\%$} & \small{ 2.7$\%$} & \small{ .044} & \small{ 5.9$\%$} & \small{ 3.6$\%$} & \small{ .057}  \\
        \midrule
		\parbox[t]{2mm}{\multirow{6}{*}{\rotatebox[origin=c]{90}{OpenEA$_{D-W-V1}$}}} 
		& MSNEA {\footnotesize {\cite{DBLP:conf/kdd/ChenL00WYC22}}}  & .238 & .452 & .31 & .254 & .465 & .326 & .318 & .514 & .385 & .432 & .601 & .490 \\
		& EVA {\footnotesize \cite{DBLP:conf/aaai/0001CRC21}}  & .570 & .801 & .653 & \underline{.575} & {.806} & {.658} & {.567} & {.797} & {.650} & .595 & .811 & .673 \\
        & MCLEA {\footnotesize {\cite{DBLP:conf/coling/LinZWSW022}}}  &
        \underline{.585} & \underline{.834} & \underline{.675} & {.574} & \underline{.824} & \underline{.663} & \underline{.581} & \underline{.813} & \underline{.665} & \underline{.655} & \underline{.848} & \underline{.726} \\
        & \CC\textbf{w/o CMMI}  
        & \CC{.640} & \CC{.879} & \CC{.727} & \CC{.644} & \CC{.882} & \CC{.730} & \CC{.667} & \CC{.891} & \CC{.749} & \CC{.722} & \CC\textbf{.908} & \CC{.790}  \\
        & \CC\textbf{UMAEA}   &
        \CC\textbf{.647} & \CC\textbf{.881} & \CC\textbf{.733} & \CC\textbf{.649} & \CC\textbf{.882} & \CC\textbf{.735} & \CC\textbf{.669} & \CC\textbf{.892} & \CC\textbf{.750} & \CC\textbf{.724} & \CC\textbf{.908} & \CC\textbf{.791} \\
        \cmidrule(lr){2-14}
        & {Improve {$\uparrow$}}  
        & \small{ 6.2$\%$} & \small{ 4.7$\%$} & \small{ .058} & \small{ 7.4$\%$} & \small{ 5.8$\%$} & \small{ .072} & \small{ 8.8$\%$} & \small{ 7.9$\%$} & \small{ .085} & \small{ 6.9$\%$} & \small{ 6.0$\%$} & \small{ .065}  \\
        \midrule
		\parbox[t]{2mm}{\multirow{6}{*}{\rotatebox[origin=c]{90}{OpenEA$_{D-W-V2}$}}} 
		& MSNEA {\footnotesize {\cite{DBLP:conf/kdd/ChenL00WYC22}}}  & .397 & .690 & .497 & .405 & .695 & .503 & .454 & .727 & .546 & .545 & .781 & .626 \\
		& EVA {\footnotesize \cite{DBLP:conf/aaai/0001CRC21}}  & \underline{.775} & .952 & .839 & \underline{.767} & {.947} & \underline{.832} & \underline{.773} & \underline{.950} & \underline{.837} & .788 & \underline{.954} & .848 \\
        & MCLEA {\footnotesize {\cite{DBLP:conf/coling/LinZWSW022}}}  &
        {.771} & \underline{.965} & \underline{.842} & {.753} & \underline{.957} & {.827} & {.757} & {.935} & {.822} & \underline{.800} & {.948} & \underline{.855} \\
        & \CC\textbf{w/o CMMI}  
        & \CC{.828} & \CC{.983} & \CC{.883} & \CC{.829} & \CC\textbf{.982} & \CC{.885} & \CC\textbf{.844} & \CC\textbf{.984} & \CC\textbf{.896} & \CC{.857} & \CC{.986} & \CC\textbf{.905}  \\
        & \CC\textbf{UMAEA}   &
        \CC\textbf{.840} & \CC\textbf{.984} & \CC\textbf{.890} & \CC\textbf{.832} & \CC\textbf{.982} & \CC\textbf{.887} & \CC\textbf{.844} & \CC\textbf{.984} & \CC\textbf{.896} & \CC\textbf{.859} & \CC\textbf{.987} & \CC\textbf{.905} \\
        \cmidrule(lr){2-14}
        & {Improve {$\uparrow$}}  
        & \small{ 6.5$\%$} & \small{ 1.9$\%$} & \small{ .048} & \small{ 6.5$\%$} & \small{ 2.5$\%$} & \small{ .055} & \small{ 7.1$\%$} & \small{ 3.4$\%$} & \small{ .059} & \small{ 5.9$\%$} & \small{ 3.3$\%$} & \small{ .050}  \\        
    \bottomrule
    \end{tabular}
    }
    \label{tab:overall}
\end{table}

\subsubsection{Implementation Details.}\label{sec:detail}
To ensure fairness, we consistently reproduce or implement all methods with the following settings:
{{(\rmnum{1})}} The hidden layer dimensions $d$ for all networks are unified into 300.  
The total epochs for baselines are set to 500 with an optional iterative training strategy applied for another 500 epochs, following \cite{DBLP:conf/coling/LinZWSW022}.
Training strategies including cosine warm-up schedule ($15\%$ steps for LR warm-up), early stopping, and gradient accumulation are adopted. The AdamW optimizer ($\beta_1=0.9$, $\beta_2=0.999$) is used, with a fixed batch size of 3500. 
{{(\rmnum{2})}} To demonstrate model stability, following \cite{DBLP:conf/ksem/ChenLWXWC20,DBLP:conf/coling/LinZWSW022}, the vision encoders $Enc_{v}$ are set to ResNet-152 \cite{DBLP:conf/cvpr/HeZRS16} on DBP15K where the vision feature dimension $d_v$ is $2048$, and set to CLIP \cite{DBLP:conf/icml/RadfordKHRGASAM21} on Multi-OpenEA with $d_v=512$.
{{(\rmnum{3})}} An alignment editing method is employed to reduce the error accumulation \cite{DBLP:conf/ijcai/SunHZQ18}.
{{(\rmnum{4})}} Following Yang et al. \cite{DBLP:conf/emnlp/YangZSLLS19}, Bag-of-Words (BoW) is selected for encoding relations ($x^r$) and attributes ($x^a$)  as fixed-length (i.e., $d_r=d_a=1000$) vectors. Specially, we firstly sort relations/attributes across KGs by frequencies in descending order. At rank $d_r$/$d_a$, we truncated or padded the list to discard the long-tail relations/attributes and obtain fixed-length all-zero vectors $x^r$ and $x^a$. For entity $e_i$: if it includes any of the top-k attributes, the corresponding position in $x_i^a$ is set to 1; if a relation of $e_i$ is among the top-k, the corresponding position in $x_i^r$ is incremented by 1.

In our UMAEA model, $\tau$ is set to 0.1 which determines how much attention the contrast loss pays to difficult negative samples. Besides, the head number $N_h$ in MHCA is set to $1$, and the training epochs are set to \{250, 50, 100\} for stage 1, 2-1, 2-2, respectively. Despite potential performance variations resulting from parameter searching, our focus remained on achieving broad applicability rather than fine-tuning for specific datasets.
During iterative training, the pipeline is repeated; but the expansion of the training set occurs exclusively in stage 1.
For MSNEA, we eliminate the attribute values for input consistency, and extend MSNEA with iterative training capability. All experiments are conducted on RTX 3090Ti GPUs.

\subsection{Overall Results} \label{sec:overall}
\subsubsection{Uncertainly Missing Modality.}
\begin{figure*}[!htbp]
  \centering
  \includegraphics[width=1.0\linewidth]{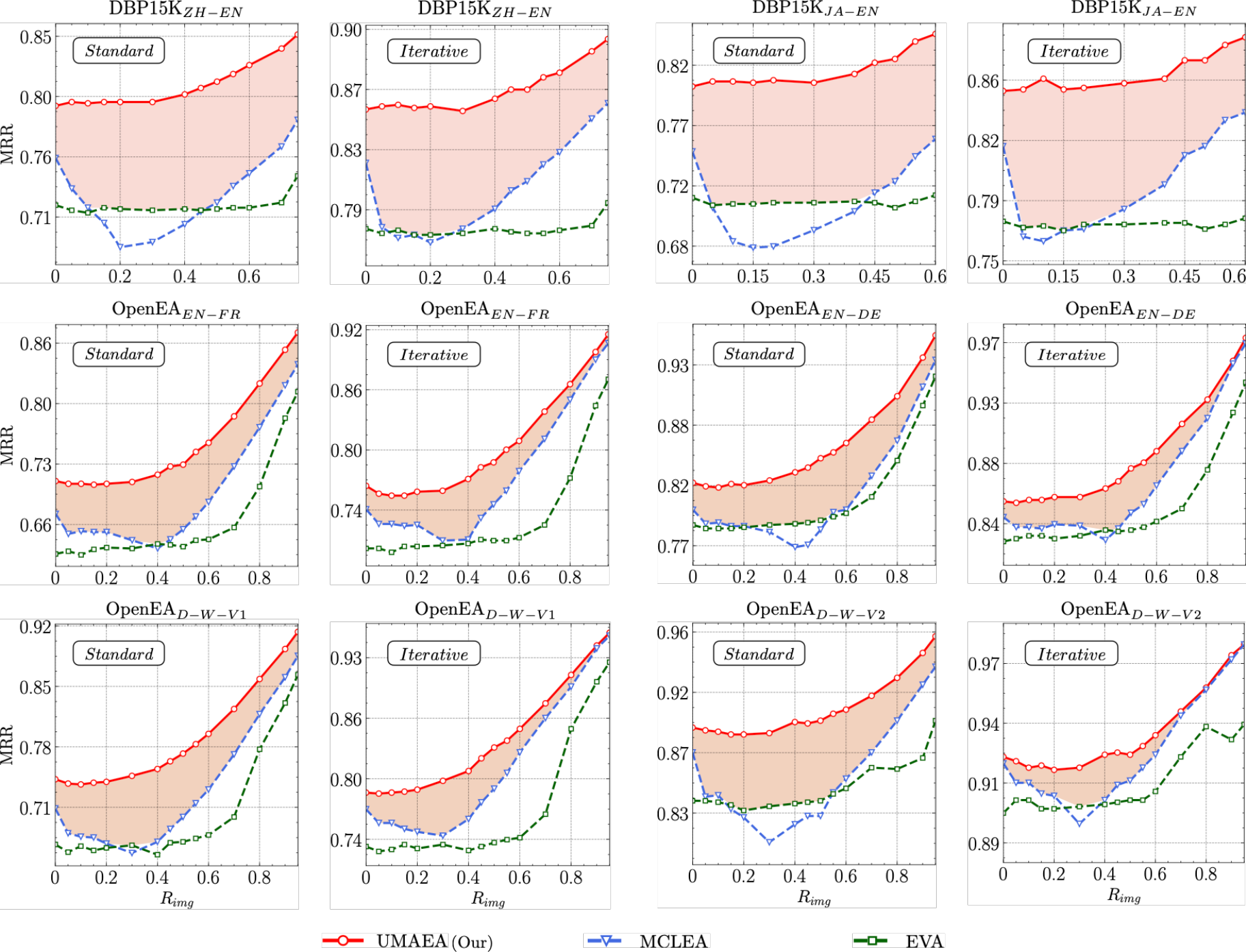}
  \caption{The overall standard (non-iterative) and iterative model performance under the setting of uncertainly missing modality with $R_{img} \in \{0.2, 0.4, 0.6\}$. The performance of DBP15K$_{FR\text{-}EN}$ are shown in Figure \ref{fig:Introcase}. }
  \label{fig:line}
\end{figure*}

Our primary experiment focuses on the model performances with varying missing modality proportions $R_{img}$.
In Table \ref{tab:overall}, we select four representative proportions: $R_{img} \in \{0.05, 0.2, 0.4, 0.6\} \times 100\%$ to simulate the degree of uncertainly missing modality that may exist in real-world scenarios, and evaluate the robustness of different models.
Our UMAEA demonstrates stable improvement on the DBP15K datasets across different $R_{img}$ values in comparison to the top-performing benchmark model:  $10.3\%$ ($R_{img}=0.05$), $11.6\%$ ($R_{img}=0.2$), $11.9\%$ ($R_{img}=0.4$), and $10.3\%$ ($R_{img}=0.6$). We note that it exhibits the most significant improvement when the $R_{img}$ lies between $20\%$ and $40\%$.
For the Multi-OpenEA datasets, our average improvement is: $5.5\%$ ($R_{img}=0.05$), $5.9\%$ ($R_{img}=0.2$), $7.3\%$ ($R_{img}=0.4$), and $6.4\%$ ($R_{img}=0.6$). Although the improvement is slightly lower than in DBP15K, the overall advantage range remains consistent, aligning with our motivation.
Besides, Figure \ref{fig:line} visualizes performance variation curves for three models. 
The overall performance trend fits the conclusions drawn in Table \ref{tab:overall}, showing that our method outperforms the baseline in terms of significant performance gap, regardless of whether iterative or non-iterative learning is employed.

Additionally, we notice a phenomenon that existing models exhibit performance oscillations (EVA) or even declines (MCLEA) at higher modality missing rates. This kind of adverse effect peaks within a particular $R_{img}^1$ range and gradually recovers and gains benefits as $R_{img}$ rises to a certain level $R_{img}^2$. In other words, when $0 \leq R_{img} \leq R_{img}^2$, the additional multi-modal data negatively impacts EA. 
This observation seems counterintuitive since providing more information leads to side effects, but it is also logical. Introducing images for half of the entities means that the remaining half may become noise, which calls for a necessary trade-off.
Under the standard (non-iterative) setting, MCLEA's $R_{img}^2$ averages $63.6\%$, which is $57.14\%$ for MSNEA and $46.43\%$ for EVA across seven datasets. Our method, augmented with the CMMI module, reaches $20.71\%$ for $R_{img}^2$. Even without CMMI, the $R_{img}^2$ of UMAEA remains at $34.29\%$. 
This implies that our method can gain benefits with fewer visual modality data in entity.
Meanwhile, UMAEA exhibits less oscillation and greater robustness than other methods, as further evidenced by the entity distribution analysis in Section \ref{sec:dist}.

We observe that our performance improvement on Multi-OpenEA is less pronounced compared to the DBP15K dataset.  This may be due to the higher image feature quality of CLIP compared to ResNet-152, which in turn diminishes the relative benefit of our model in addressing feature ambiguity. Additionally, as the appendix shows, these datasets have fewer relation and attribute types, allowing for better feature training with comparable data sizes (with a fixed 1000-word bag size, long tail effects are minimized) which partially compensates for missing image modalities.  This finding can also explain why, as seen in Figure \ref{fig:line}, our model's performance improvement decreases as $R_{img}$ increases, and our enhancement in the dense graph (D-W-V2) is slightly less pronounced than in the sparse graph (D-W-V1) which has richer graph structure information.

\subsubsection{Complete Modality.}
We also evaluate our model on the standard multi-modal DBP15K \cite{DBLP:conf/aaai/0001CRC21} dataset, achieving satisfactory results with or without the visual modality (w/o IMG), as shown in Table \ref{tab:std-1}. It is noteworthy that the DBP15K dataset only has part of the entities with images attached (e.g., $78.29\%$ in DBP15K$_{ZH\text{-}EN}$, $70.32\%$ in DBP15K$_{FR\text{-}EN}$, and $67.58\%$ in DBP15K$_{JA\text{-}EN}$), which is inherent to the DBPedia database.
To further showcase our method's adaptability, in Table \ref{tab:std-2}, we evaluate it on the standard Multi-OpenEA dataset with $100\%$ image data attached, demonstrating that our method can be superior in the (MM)EA task against the potentially ambiguous modality information.
\begin{table}[!htbp]
    \centering
	\tabcolsep=0.3cm
    \renewcommand\arraystretch{1.0}
    \caption{Non-iterative (Non-iter.) and iterative (Iter.) results on three multi-modal DPB15K \cite{DBLP:conf/semweb/SunHL17} datasets, where `` * '' refers to involving the visual information for EA. 
    }
    \resizebox{0.87\linewidth}{!}{
    \begin{tabular}{@{}l|l|ccc|ccc|ccc}
        \toprule
        & \multirow{2}*{\makebox[2cm][c]{Models}} & \multicolumn{3}{c|}{DBP15K$_{ZH-EN}$} & \multicolumn{3}{c|}{DBP15K$_{JA-EN}$} & \multicolumn{3}{c}{DBP15K$_{FR-EN}$} \\
        & & {\scriptsize H@1} & {\scriptsize H@10} & {\scriptsize MRR} & {\scriptsize H@1} & {\scriptsize H@10} & {\scriptsize MRR} & {\scriptsize H@1} & {\scriptsize H@10} & {\scriptsize MRR} \\
        \midrule
        \parbox[t]{2mm}{\multirow{8}{*}{\rotatebox[origin=c]{90}{Non-iter.}}} 
        & AlignEA {\footnotesize \cite{DBLP:conf/ijcai/SunHZQ18}} & 
        .472 & .792 & .581 & .448 & .789 & .563 & .481 & .824 & .599 \\
        & KECG {\footnotesize {\cite{DBLP:conf/emnlp/LiCHSLC19}}} &
        .478 & .835 & .598 & .490 & .844 & .610 & .486 & .851 & .610 \\
        & MUGNN {\footnotesize \cite{DBLP:conf/acl/CaoLLLLC19}} &
        .494 & .844 & .611 &  .501 & .857 & .621 & .495 & .870 & .621 \\
        & AliNet {\footnotesize {\cite{DBLP:conf/aaai/SunW0CDZQ20}}} &
        .539 & .826 & .628 & .549 & .831 & .645 & .552 & .852 & .657 \\
        & MSNEA* {\footnotesize {\cite{DBLP:conf/kdd/ChenL00WYC22}}} & .609 & .831 & .685 & .541 & .776 & .620 & .557 & .820 & .643 \\
        & EVA* {\footnotesize \cite{DBLP:conf/aaai/0001CRC21}} &
        {.683} & {.906} & {.762} & {.669} & {.904} & {.752} & {.686} & \underline{.928} & {.771} \\
        & MCLEA* {\footnotesize {\cite{DBLP:conf/coling/LinZWSW022}}}  &
        \underline{.726} & \underline{.922} & \underline{.796} & \underline{.719} & \underline{.915} & \underline{.789} & \underline{.719} & {.918} & \underline{.792} \\
        & \CC\textbf{UMAEA*}  &
        \CC\textbf{.800} & \CC\textbf{.962} & \CC\textbf{.860} & \CC\textbf{.801} & \CC\textbf{.967} & \CC\textbf{.862} & \CC\textbf{.818} & \CC\textbf{.973} & \CC\textbf{.877} \\
        & \CC\textbf{~~ w/o IMG}  &
        \CC\textbf{.718} & \CC\textbf{.930} & \CC\textbf{.797} & \CC\textbf{.723} & \CC\textbf{.941} & \CC\textbf{.803} & \CC\textbf{.748} & \CC\textbf{.956} & \CC\textbf{.826} \\
        \midrule
        \parbox[t]{2mm}{\multirow{6}{*}{\rotatebox[origin=c]{90}{Iter.}}} 
        & BootEA {\footnotesize \cite{DBLP:conf/ijcai/SunHZQ18}} & 
        .629 & .847 & .703 & .622 & .854 & .701 & .653 & .874 & .731 \\
        & NAEA {\footnotesize {\cite{DBLP:conf/ijcai/ZhuZ0TG19}}} &
        .650 & .867 & .720 & .641 & .873 & .718 & .673 & .894 & .752 \\
        & MSNEA* {\footnotesize {\cite{DBLP:conf/kdd/ChenL00WYC22}}} & .648 & .881 & .728 & .557 & .804 & .643 & .583 & .848 & .672 \\
        & EVA* {\footnotesize \cite{DBLP:conf/aaai/0001CRC21}} &
        {.750} & {.912} & {.810} & {.741} & {.921} & {.807} & {.765} & {.944} & {.831} \\
        & MCLEA* {\footnotesize {\cite{DBLP:conf/coling/LinZWSW022}}}  &
        \underline{.811} & \underline{.957} & \underline{.865} & \underline{.805} & \underline{.958} & \underline{.863} & \underline{.808} & \underline{.963} & \underline{.867} \\
        & \CC\textbf{UMAEA*}  &
        \CC\textbf{.856} & \CC\textbf{.974} & \CC\textbf{.900} & \CC\textbf{.857} & \CC\textbf{.980} & \CC\textbf{.904} & \CC\textbf{.873} & \CC\textbf{.988} & \CC\textbf{.917}   \\
        & \CC\textbf{~~ w/o IMG}  &
        \CC\textbf{.793} & \CC\textbf{.952} & \CC\textbf{.852} & \CC\textbf{.794} & \CC\textbf{.960} & \CC\textbf{.857} & \CC\textbf{.820} & \CC\textbf{.976} & \CC\textbf{.880} \\
        \bottomrule
    \end{tabular}
    }
    \label{tab:std-1}
\end{table}

\begin{table}[!htbp]
    \centering
	\tabcolsep=0.3cm
    \renewcommand\arraystretch{1.0}
    \caption{Non-iterative (Non-iter.) and iterative (Iter.) results on four standard Multi-OpenEA \cite{DBLP:journals/corr/abs-2302-08774} datasets with $R_{img}=1.0$.  
    }
    \resizebox{0.98\linewidth}{!}{
    \begin{tabular}{@{}l|l|ccc|ccc|ccc|ccc}
        \toprule
        & \multirow{2}*{\makebox[2cm][c]{Models}} & \multicolumn{3}{c|}{OpenEA$_{EN-FR}$} & \multicolumn{3}{c|}{OpenEA$_{EN-DE}$} & \multicolumn{3}{c|}{OpenEA$_{D-W-V1}$} & \multicolumn{3}{c}{OpenEA$_{D-W-V2}$} \\
        & & {\scriptsize H@1} & {\scriptsize H@10} & {\scriptsize MRR} & {\scriptsize H@1} & {\scriptsize H@10} & {\scriptsize MRR} & {\scriptsize H@1} & {\scriptsize H@10} & {\scriptsize MRR} & {\scriptsize H@1} & {\scriptsize H@10} & {\scriptsize MRR} \\
        \midrule
        \parbox[t]{2mm}{\multirow{4}{*}{\rotatebox[origin=c]{90}{Non-iter.}}} 
        & MSNEA* {\footnotesize {\cite{DBLP:conf/kdd/ChenL00WYC22}}} 
        & .692 & .813 & .734 & .753 & .895 & .804 & .800 & .874 & .826 & .838 & .940 & .873 \\
        & EVA* {\footnotesize \cite{DBLP:conf/aaai/0001CRC21}} 
        & {.785} & {.932} & {.836} & {.922} & {.983} & {.945} & {.858} & {.946} & {.891} & .890 & .981 & .922 \\
        & MCLEA* {\footnotesize {\cite{DBLP:conf/coling/LinZWSW022}}}  
        & \underline{.819} & \underline{.943} & \underline{.864} & \underline{.939} & \underline{.988} & \underline{.957} & \underline{.881} & \underline{.955} & \underline{.908} & \underline{.928} & \underline{.983} & \underline{.949} \\
        & \CC\textbf{UMAEA*}  
        & \CC\textbf{.848} & \CC\textbf{.966} & \CC\textbf{.891} & \CC\textbf{.956} & \CC\textbf{.994} & \CC\textbf{.971} & \CC\textbf{.904} & \CC\textbf{.971} & \CC\textbf{.930} & \CC\textbf{.948} & \CC\textbf{.996} & \CC\textbf{.967} \\
        \midrule
        \parbox[t]{2mm}{\multirow{4}{*}{\rotatebox[origin=c]{90}{Iter.}}}
        & MSNEA* {\footnotesize {\cite{DBLP:conf/kdd/ChenL00WYC22}}} 
        & .699 & .823 & .742 & .788 & .917 & .835 & .809 & .885 & .836 & .862 & .954 & .894 \\
        & EVA* {\footnotesize \cite{DBLP:conf/aaai/0001CRC21}} 
        & {.849} & {.974} & {.896} & {.956} & {.985} & {.968} & {.915} & {.986} & {.942} & .925 & .996 & .951 \\
        & MCLEA* {\footnotesize {\cite{DBLP:conf/coling/LinZWSW022}}}  
        & \underline{.888} & \underline{.979} & \underline{.924} & \underline{.969} & \underline{.993} & \underline{.979} & \underline{.944} & \underline{.989} & \underline{.963} & \underline{.969} & \underline{.997} & \underline{.982} \\
        & \CC\textbf{UMAEA*}  
        & \CC\textbf{.895} & \CC\textbf{.987} & \CC\textbf{.931} & \CC\textbf{.974} & \CC\textbf{998} & \CC\textbf{.984} & \CC\textbf{.945} & \CC\textbf{.994} & \CC\textbf{.965} & \CC\textbf{.973} & \CC\textbf{.999} & \CC\textbf{.984} \\
        \bottomrule
    \end{tabular}
    }
    \label{tab:std-2}
\end{table}


\begin{figure*}[!htbp]
  \centering
  \includegraphics[width=0.98\linewidth]{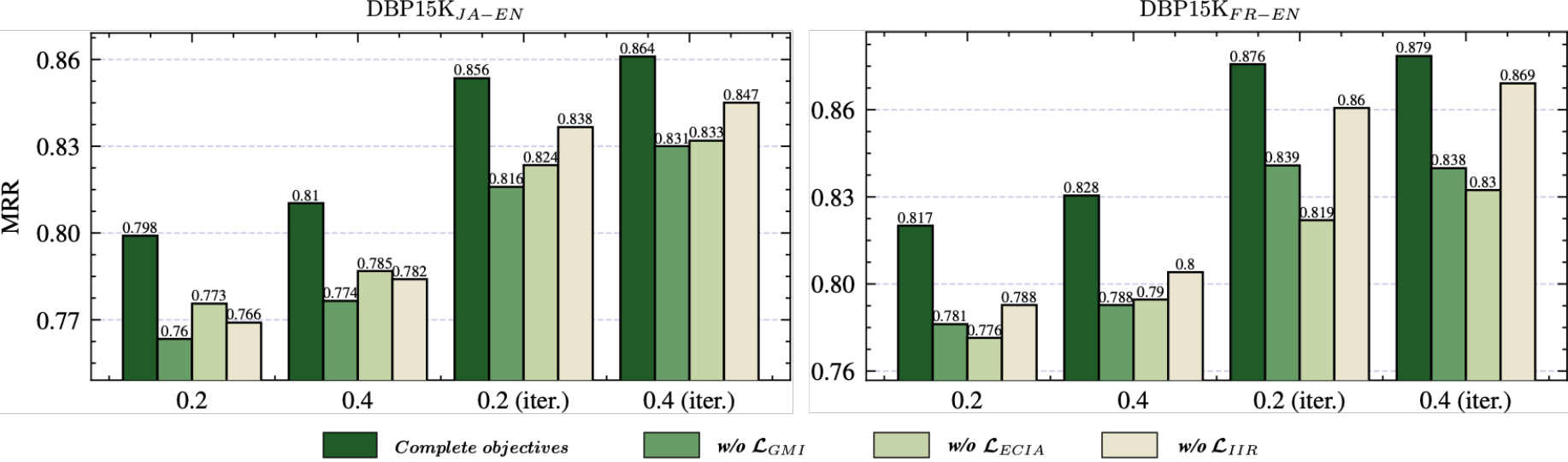}
  \caption{The component analysis of UMAEA (w/o CMMI), where the scales on the horizontal axis represent $R_{img} \in \{0.2, 0.4\}$ and ``iter.'' represents the model performance on iterative setting.}
  \label{fig:ablation}
\end{figure*}
\subsection{Details Analysis} \label{sec:analysis}
\subsubsection{{Component Analysis.}}
We further analyze the impact of each training objective on our model's performance in Figure \ref{fig:ablation}, where the absence of any objective results in varying performance degradation. 
As mentioned in Section \ref{sec:iir}, IIR serves as an enhancement for ECIA, and its influence is comparatively less significant than that of $\mathcal{L}_{GMI}$ and $\mathcal{L}_{ECIA}$.
The CMMI module's influence is detailed in Table \ref{tab:overall}, where it becomes more significant when $R_{img}$ is low. CMMI's primary function is to mitigate noise in the missing modalities,  facilitating efficient learning at high noise levels and minimizing the noise to existing information.

\subsubsection{{Efficiency Analysis.}}
Concurrently, we briefly compare the relationship between model parameter size, training time, and performance. Our model improves the  performance with only a minor increase in parameters and time consumption. 
This indicates that in many cases, our method can directly substitute these models with minimal additional overhead.
While there is potential for enhancing UMAEA's efficiency, we view this as a direction for future research.

\begin{table}[!htbp]
    \centering
	\tabcolsep=0.3cm
    \renewcommand\arraystretch{1.0}
    \caption{Efficiency Analysis. Non-iterative model performance on three datasets with $R_{img}=0.4$, where ``Para.'' refers to the number of learnable parameters and ``Time'' refers to the total time required for model to reach the optimal performance.
    }
    \resizebox{1.0\linewidth}{!}{
    \begin{tabular}{l|ccc|ccc|ccc}
        \toprule
         \multirow{2}*{\makebox[2cm][c]{Models}} & \multicolumn{3}{c|}{DBP15K$_{JA-EN}$} & \multicolumn{3}{c|}{DBP15K$_{FR-EN}$} & \multicolumn{3}{c}{OpenEA$_{EN-FR}$} \\
        & {\scriptsize Para. (M) } & {\scriptsize Time (Min) } & {\scriptsize MRR } & {\scriptsize Para. (M) } & {\scriptsize Time (Min) } & {\scriptsize MRR } & {\scriptsize Para. (M) } & {\scriptsize Time (Min) } & {\scriptsize MRR } \\
        \midrule
         EVA* {\footnotesize \cite{DBLP:conf/aaai/0001CRC21}} 
        & 13.27 & 30.9 & .711 & 13.29 & 30.8 & .721 & 9.81 & 17.8 & .642   \\
         MCLEA* {\footnotesize {\cite{DBLP:conf/coling/LinZWSW022}}}  
        & 13.22 & 15.3 & .703 & 13.24 & 15.7 & .702 & 9.75 & 19.5 & .637   \\
        \CC{w/o CMMI}  
        & \CC{13.82} & \CC{30.2} & \CC{.810} & \CC{13.83} & \CC{28.8} & \CC{.828} & \CC{10.35} & \CC{17.9} & \CC{.715}   \\
        \CC{UMAEA}  
        & \CC{14.72} & \CC{33.4} & \CC{.813} & \CC{14.74} & \CC{32.7} & \CC{.838} & \CC{11.26} & \CC{23.1} & \CC{.718} \\
        \bottomrule
    \end{tabular}
    }
    \label{tab:overall-iter-2}
\end{table}

\begin{figure*}[!htbp]
  \centering
  \includegraphics[width=0.97\linewidth]{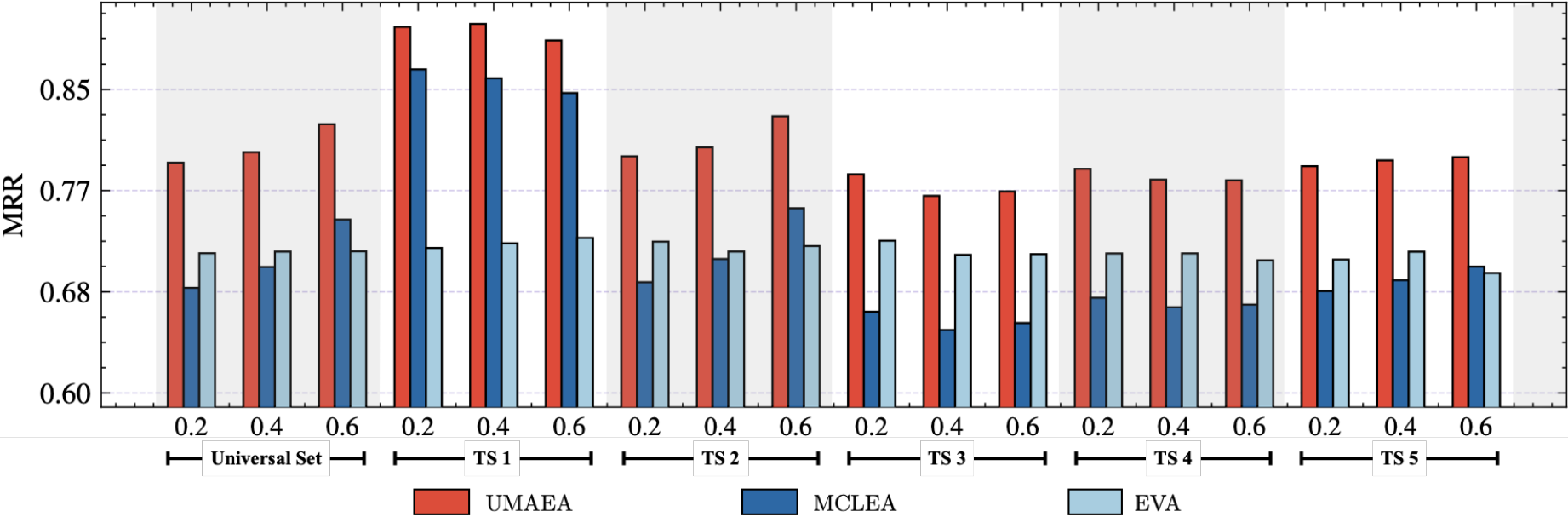}
  \caption {EA prediction distribution analysis on DBP15K$_{ZH-EN}$ (non-iterative), with
  $R_{img} \in \{0.2, 0.4, 0.6\}$. ``TS'' denotes the testing set, where:
  		TS 1 (both entities in an alignment pair have images);
		TS 2 (at least one entity in an alignment pair has images);
		TS 3 (only one entity in an alignment pair has images);
            TS 4 (at least one entity in an alignment pair loss images);
		TS 5 (neither entity in an alignment pair has images).
  }
  \label{fig:case}
\end{figure*}
\subsubsection{Entity Distribution Analysis.} \label{sec:dist}
To further evaluate the robustness of our method,
we analyze the model's prediction performance under different distributions of entity's visual modality.  Concretely, we compare five testing sets under $R_{img} \in  \{0.2, 0.4, 0.6\}$ with details presented in Figure \ref{fig:case},
where we exclude the CMMI module during the comparison.
We observe that EVA's performance is generally stable but underperforms when visual modality is  complete (TS 1), suggesting its overfitting to modality noise in the training stage.
In contrast, MCLEA exhibits more extreme performance fluctuations, performing worse than EVA does when there's incomplete visual information within the entity pairs (TS 2, 3, 4, 5).
Our superior performance reflects the intuition that the optimal performance occurs in TS 1, with tolerable fluctuations in other scenarios.

\section{Conclusion}
In this work, we discussed the challenges and limitations of existing MMEA methods in dealing with modality incompleteness and visual ambiguity. Our analysis revealed that certain models overfit to modality noise and suffer from oscillating or declining performance at high modality missing rates, emphasizing the need for a more robust approach. Thus, we introduced UMAEA which introduces multi-scale modality hybrid and circularly missing modality imagination to tackle this problem, performing well across all benchmarks. 
There remain opportunities for future research, such as evaluating our techniques for the incompleteness of other modalities (e.g., attribute), and investigating effective techniques to  utilize more detailed visual contents for MMEA.

\subsubsection{Acknowledgments.}
This work was supported by the National Natural Science Foundation of China (NSFCU19B2027/NSFC91846204), joint project DH-2022ZY0012 from Donghai Lab, and the EPSRC project ConCur (EP/V050869/1).

%
%
%

%

 \bibliographystyle{splncs04}
 \bibliography{mybibliography}
\appendix
\clearpage

\appendix
\section{Appendix}
\begin{table}[!htbp]
    \centering
    \vspace{-0.1cm}
    \caption{Statistics for original datasets, where ``EA pairs'' refers to the pre-aligned entity pairs. Note that not all entities have the associated images or the equivalent counterparts in the other KG. For dataset \{ EN-FR-15K, EN-DE-15K, D-W-15K-V1, and D-W-15K-V2 \} in Multi-OpenEA, we omit the ``15K'' suffix to unify the description throughout this paper.}
    \label{tab:dataset}
    \vspace{-3pt}
    \renewcommand\arraystretch{1.0}
    \resizebox{1.\linewidth}{!}{
    \begin{tabular}{@{}l|c|cccccccc@{}}
        \toprule
        \makebox[2.5cm][c]{Dataset} & KG & \# Ent. & \# Rel. & \# Attr. & \# Rel. Triples & \# Attr. Triples & \# Image & \# EA pairs \\
        \midrule
        \multirow{2}*{DBP15K$_{ZH\text{-}EN}$} & ZH {\footnotesize (Chinese)} & 19,388 & 1,701 & 8,111 & 70,414 & 248,035 & 15,912 & \multirow{2}*{15,000} \\
        & EN {\footnotesize (English)} & 19,572 & 1,323 & 7,173 & 95,142 & 343,218 & 14,125 \\
        \midrule
        \multirow{2}*{DBP15K$_{JA\text{-}EN}$} & JA {\footnotesize (Japanese)} & 19,814 & 1,299 & 5,882 & 77,214 & 248,991 & 12,739 & \multirow{2}*{15,000} \\
        & EN {\footnotesize (English)} & 19,780 & 1,153 & 6,066 & 93,484 & 320,616 & 13,741 \\
        \midrule
        \multirow{2}*{DBP15K$_{FR\text{-}EN}$} & FR {\footnotesize (French)} & 19,661 & 903 & 4,547 & 105,998 & 273,825 & 14,174 & \multirow{2}*{15,000} \\
        & EN {\footnotesize (English)} & 19,993 & 1,208 & 6,422 & 115,722 & 351,094 & 13,858 \\
        \midrule
        \multirow{2}*{OpenEA$_{EN\text{-}FR}$} & EN {\footnotesize (English)} & 15,000 & 267 & 308 & 47,334 & 73,121 & 15,000 & \multirow{2}*{15,000} \\
        & FR {\footnotesize (French)} & 15,000 & 210 & 404 & 40,864 & 67,167 & 15,000 \\
        \midrule
        \multirow{2}*{OpenEA$_{EN\text{-}DE}$} & EN {\footnotesize (English)} & 15,000 & 215 & 286 & 47,676 & 83,755 & 15,000 & \multirow{2}*{15,000} \\
        & DE (German) & 15,000 & 131 & 194 & 50,419 & 156,150 & 15,000 \\
                \midrule
        \multirow{2}*{OpenEA$_{D\text{-}W\text{-}V1}$} & DBpedia & 15,000 & 248 & 342 & 38,265 & 68,258 & 15,000 & \multirow{2}*{15,000} \\
        & Wikidata & 15,000 & 169 & 649 & 42,746 & 138,246 & 15,000 \\
        \midrule
        \multirow{2}*{OpenEA$_{D\text{-}W\text{-}V2}$} & DBpedia & 15,000 & 167 & 175 & 73,983 & 66,813 & 15,000 & \multirow{2}*{15,000} \\
        & Wikidata & 15,000 & 121 & 457 & 83,365 & 175,686 & 15,000 \\
        \bottomrule
    \end{tabular}
    }
    \vspace{-0.6cm}
\end{table}

\begin{table}[!htbp]
  \centering
\centering
{
\vspace{-2pt}
\caption{The proportion $R_{img}$ of entities containing images for each dataset in our setting, with ``\texttt{STD}'' refers to the standard $R_{img}$ in raw datasets.
}
\label{tab:split}
}
\vspace{-2pt}
\resizebox{1.0\linewidth}{!}{
\begin{tabular}{@{}l|l}
\hline
 & \\ [-2ex]
\makebox[2.5cm][c]{Dataset} & \makebox[14cm][c]{$R_{img}$} \\
 & \\ [-2ex]
\hline
DBP15K$_{ZH\text{-}EN}$
&\texttt{\small ~0.05, 0.1, 0.15, 0.2, 0.3, 0.4, 0.45, 0.5, 0.55, 0.6, 0.7, 0.75, 0.7829~(STD)}\\
& \\ [-2ex]
\hline
 & \\ [-2ex]
DBP15K$_{JA\text{-}EN}$
&\texttt{\small ~0.05, 0.1, 0.15, 0.2, 0.3, 0.4, 0.45, 0.5, 0.55, 0.6, 0.7, 0.7032~(STD)}\\
& \\ [-2ex]
\hline
 & \\ [-2ex]
DBP15K$_{FR\text{-}EN}$
&\texttt{\small ~0.05, 0.1, 0.15, 0.2, 0.3, 0.4, 0.45, 0.5, 0.55, 0.6, 0.6758  (STD) }\\
& \\ [-2ex]
\hline
 & \\ [-2ex]
OpenEA$_{EN\text{-}FR}$
&\texttt{\small ~0.05, 0.1, 0.15, 0.2, 0.3, 0.4, 0.45, 0.5, 0.55, 0.6, 0.7, 0.8, 0.9, 0.95, 1.0~(STD)}\\ 
& \\ [-2ex]
\hline
 & \\ [-2ex]
OpenEA$_{EN\text{-}DE}$
&\texttt{\small ~0.05, 0.1, 0.15, 0.2, 0.3, 0.4, 0.45, 0.5, 0.55, 0.6, 0.7, 0.8, 0.9, 0.95, 1.0~(STD)}\\ 
& \\ [-2ex]
\hline
 & \\ [-2ex]
OpenEA$_{D\text{-}W\text{-}V1}$
&\texttt{\small ~0.05, 0.1, 0.15, 0.2, 0.3, 0.4, 0.45, 0.5, 0.55, 0.6, 0.7, 0.8, 0.9, 0.95, 1.0~(STD)}\\ 
& \\ [-2ex]
\hline
 & \\ [-2ex]
OpenEA$_{D\text{-}W\text{-}V2}$
&\texttt{\small ~0.05, 0.1, 0.15, 0.2, 0.3, 0.4, 0.45, 0.5, 0.55, 0.6, 0.7, 0.8, 0.9, 0.95, 1.0~(STD)}\\ 
[-2ex] \\
\hline
\end{tabular}}
\vspace{-0.6cm}
\end{table}

\subsection{Dataset Statistics}
Our detailed dataset statistics are presented in Table \ref{tab:dataset}.
A set of pre-aligned entity pairs is offered for guidance,  
which is proportionally split into a training set (seed alignments $\mathcal{S}$) and a testing set $\mathcal{S}_{te}$ based on the given seed alignment ratio ($R_{sa}$).
Notably,   each entity in the four Multi-OpenEA benchmark \cite{DBLP:journals/corr/abs-2302-08774} is initially associated with three images obtained from the Google search engine. In this study, we select the highest-ranked image, which is the first one, to serve as the visual information for the entity.
 The details for 97 data splits are contained in Table \ref{tab:split}, and the complete data for benchmark is accessible at {\color{blue}\url{https://github.com/zjukg/UMAEA}}.

\subsection{Supplementary for Experiments}
\begin{figure*}[!htbp]
  \centering
  \vspace{-0.2cm}
  \includegraphics[width=1.0\linewidth]{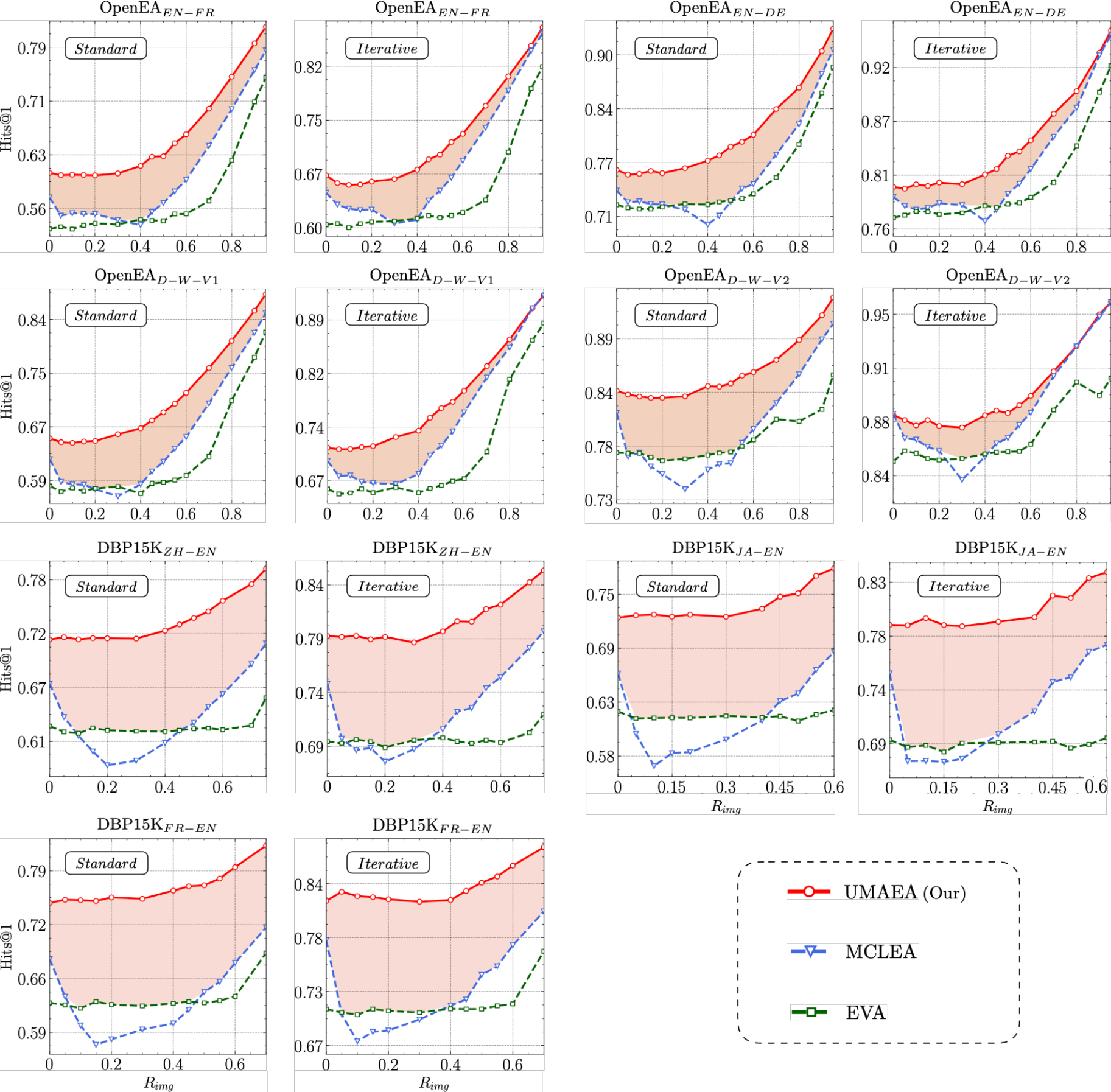}
  \vspace{-0.1cm}
  \caption{The overall standard (non-iterative) and iterative model performance (\textbf{Hit$@$1}) under the setting of uncertainly missing modality with $R_{img} \in \{0.2, 0.4, 0.6\}$. }
 \vspace{-4pt}
  \label{fig:appline}
\vspace{-0.2cm}
\end{figure*}
\subsubsection{{Experiment Settings.}}
Those attribute triples $<$\textit{entity}, \textit{attribute}, \textit{value}$>$ in KGs  have been researched in many previous EA works \cite{DBLP:conf/aaai/TrisedyaQZ19,DBLP:conf/emnlp/LiuCPLC20,DBLP:conf/ijcai/Tang0C00L20,DBLP:conf/kdd/ChenL00WYC22,DBLP:conf/icde/ZhongZFD22}.
Nevertheless, in order to focus on our key subject,
we do not utilize the contents of \textit{value} parts in this work which are mainly string formats like specific date, land area or coordinate position.
Furthermore, in order to concentrate on uncertainly missing visual modality, we exclude surface-related information such as the name of entity, relation, and attribute. Our approach primarily utilizes information derived from the type of entity and relationship, as well as structure of the graph and the image data, which is inherited from previous works \cite{DBLP:conf/aaai/0001CRC21,DBLP:conf/kdd/ChenL00WYC22,DBLP:conf/coling/LinZWSW022}. Each entity is associated with multiple attributes and either $0$ or $1$ image. We achieve this association through id/index sharing, following previous works \cite{DBLP:conf/kdd/ChenL00WYC22,DBLP:journals/corr/abs-2302-08774,DBLP:conf/coling/LinZWSW022,DBLP:conf/aaai/0001CRC21}, rather than explicitly defining triples. For example, Wang et al. \cite{DBLP:journals/bdr/WangWQZ20} incorporate images as entities through the introduction of a specific \textit{Imageof} relation, allowing for a more formal structure and organization of the KG.

 Regarding the loss trade-off for multi-task learning, we attempted to use the Automatic Weighted Loss (AWL) technique \cite{DBLP:conf/cvpr/KendallGC18} to dynamically assign weights to different training objectives. However, we found that directly summing the losses after scaling resulted in similar performance ($\pm$ 0.3$\%$ in hit$@$1) compared to using AWL. Hence, we omitted this empirical study in the paper.

Regarding $R_{img}^2$ for MCLEA, as mentioned before, the adverse effect gradually recovers and gains benefits as $R_{img}$ rises to a certain level $R_{img}^2$. Here, $R_{img}^2$ represents the minimum observed $R_{img}$ at which the model's performance surpasses that without visual information ($R_{img}$=0). For MCLEA, we calculate as follows:: 
$[0.7(\text{ZH-EN})+0.7(\text{FR-EN})+0.6(\text{JA-EN})+0.55(\text{D-W-v1})+0.7(\text{D-W-v2})+0.6(\text{EN-DE})+0.6(\text{EN-FR})]/7\times100\%=63.6\%$

\vspace{-0.3cm}
\subsubsection{{Additional Experiments.}}
\begin{figure*}[!htbp]
  \centering
  \includegraphics[width=0.98\linewidth]{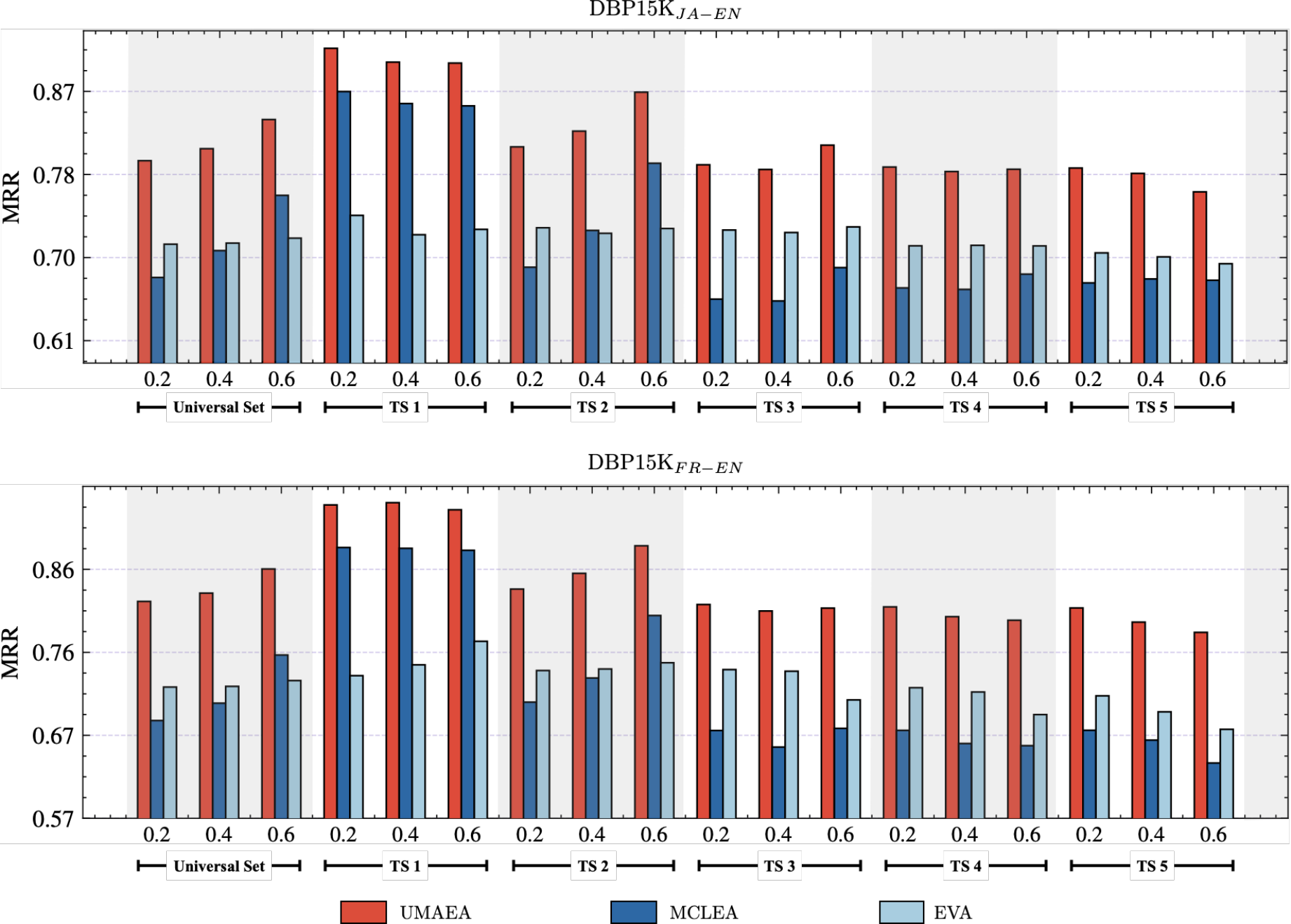}
  \caption {EA prediction distribution analysis on DBP15K$_{JA-EN}$ and DBP15K$_{FR-EN}$ (non-iterative), with $R_{img} \in \{0.2, 0.4, 0.6\}$. ``TS X'' denotes the X part of the testing set, where:
  		TS 1 (both entities in an alignment pair have images);
		TS 2 (at least one entity in an alignment pair has images);
		TS 3 (only one entity in an alignment pair has images);
            TS 4 (at least one entity in an alignment pair loss images);
		TS 5 (neither entity in an alignment pair has images).
  }
\vspace{-0.3cm}
  \label{fig:appcase}
  \vspace{-10pt}
\end{figure*}

In this section, we provide the remaining benchmark results. 
As a supplement to Figure 5, we offer a performance comparison of models for DBP15K$_{JA-EN}$ and DBP15K$_{FR-EN}$ under different testing sets, as shown in Figure \ref{fig:appcase}, which is consistent to DBP15K$_{ZH-EN}$.

Table \ref{tab:appdbp} and Table \ref{tab:appoea} present the model performance when they are applied to typical EA tasks excluding the influence of visual modality, which obviates the need for the CMMI module during training. 
The results show that our model achieved superior performance in non-multimodal EA tasks, indicating that UMAEA can even effectively mitigate the impact of information imbalance issues arising from attribute, relation, and graph structure during model training. 
Furthermore, we provide the performance curves under the Hit$@$1 metric, as illustrated in Figure \ref{fig:appline}, where the general trend in performance change closely resembles that observed under the MRR metric (Figure \ref{fig:line}).

\begin{table}[!htbp]
    \centering
	\tabcolsep=0.3cm
	\vspace{-0.2cm}
    \renewcommand\arraystretch{1.0}
    \caption{Non-iterative (Non-iter.) and iterative (Iter.) results on three standard DPB15K \cite{DBLP:conf/semweb/SunHL17} datasets with $R_{sa}=0.3$ without the visual modality ($R_{img}=0$). 
    }
    \resizebox{0.84\linewidth}{!}{
    \begin{tabular}{@{}l|l|ccc|ccc|ccc}
        \toprule
        & \multirow{2}*{\makebox[2cm][c]{Models}} & \multicolumn{3}{c|}{DBP15K$_{ZH-EN}$} & \multicolumn{3}{c|}{DBP15K$_{JA-EN}$} & \multicolumn{3}{c}{DBP15K$_{FR-EN}$} \\
        & & {\scriptsize H@1} & {\scriptsize H@10} & {\scriptsize MRR} & {\scriptsize H@1} & {\scriptsize H@10} & {\scriptsize MRR} & {\scriptsize H@1} & {\scriptsize H@10} & {\scriptsize MRR} \\
        \midrule
        \parbox[t]{2mm}{\multirow{5}{*}{\rotatebox[origin=c]{90}{Non-iter.}}} 
        & MSNEA {\footnotesize {\cite{DBLP:conf/kdd/ChenL00WYC22}}} & .503 & .795 & .602 & .395 & .715 & .504 & .472 & .820 & .593 \\
        & EVA {\footnotesize \cite{DBLP:conf/aaai/0001CRC21}} &
        {.629} & {.882} & {.719} & {.627} & {.879} & {.714} & {.626} & {.896} & {.722} \\
        & MCLEA {\footnotesize {\cite{DBLP:conf/coling/LinZWSW022}}}  &
        {.672} & {.907} & {.756} & {.663} & {.904} & {.751} & {.679} & {.923} & {.769} \\
         & MEAformer {\footnotesize {\cite{chen2023meaformer}}}  &
        \underline{.708} & \underline{.925} & \underline{.787} & \underline{.699} & \underline{.934} & \underline{.785} & \underline{.722} & \underline{.947} & \underline{.805} \\
        & \CC\textbf{UMAEA}  &
        \CC\textbf{.718} & \CC\textbf{.930} & \CC\textbf{.797} & \CC\textbf{.723} & \CC\textbf{.941} & \CC\textbf{.803} & \CC\textbf{.748} & \CC\textbf{.956} & \CC\textbf{.826} \\
        \midrule
        \parbox[t]{2mm}{\multirow{5}{*}{\rotatebox[origin=c]{90}{Iter.}}} 
        & MSNEA {\footnotesize {\cite{DBLP:conf/kdd/ChenL00WYC22}}} & .545 & .850 & .648 & .451 & .788 & .567 & .531 & .872 & .648 \\
        & EVA {\footnotesize \cite{DBLP:conf/aaai/0001CRC21}} &
        {.696} & {.907} & {.774} & {.695} & {.908} & {.772} & {.708} & {.930} & {.790} \\
        & MCLEA {\footnotesize {\cite{DBLP:conf/coling/LinZWSW022}}}  &
        {.749} & {.933} & {.817} & {.752} & {.935} & {.821} & {.779} & {.955} & {.847} \\
        & MEAformer {\footnotesize {\cite{chen2023meaformer}}}  &
        \underline{.775} & \underline{.940} & \underline{.837} & \underline{.761} & \underline{.950} & \underline{.831} & \underline{.785} & \underline{.963} & \underline{.852} \\
        & \CC\textbf{UMAEA}  &
        \CC\textbf{.793} & \CC\textbf{.952} & \CC\textbf{.852} & \CC\textbf{.794} & \CC\textbf{.960} & \CC\textbf{.857} & \CC\textbf{.820} & \CC\textbf{.976} & \CC\textbf{.880} \\
        \bottomrule
    \end{tabular}
    }
    \label{tab:appdbp}
	\vspace{-0.3cm}
\end{table}

\begin{table}[!htbp]
    \centering
	\tabcolsep=0.3cm
    \renewcommand\arraystretch{1.0}
    \caption{Non-iterative (Non-iter.) and iterative (Iter.) results on four standard OpenEA \cite{DBLP:journals/pvldb/SunZHWCAL20} datasets  with $R_{sa}=0.2$ without the visual modality ($R_{img}=0$).  
    }
    \vspace{-1pt}
    \resizebox{0.98\linewidth}{!}{
    \begin{tabular}{@{}l|l|ccc|ccc|ccc|ccc}
        \toprule
        & \multirow{2}*{\makebox[2cm][c]{Models}} & \multicolumn{3}{c|}{OpenEA$_{EN-FR}$} & \multicolumn{3}{c|}{OpenEA$_{EN-DE}$} & \multicolumn{3}{c|}{OpenEA$_{D-W-V1}$} & \multicolumn{3}{c}{OpenEA$_{D-W-V2}$} \\
        & & {\scriptsize H@1} & {\scriptsize H@10} & {\scriptsize MRR} & {\scriptsize H@1} & {\scriptsize H@10} & {\scriptsize MRR} & {\scriptsize H@1} & {\scriptsize H@10} & {\scriptsize MRR} & {\scriptsize H@1} & {\scriptsize H@10} & {\scriptsize MRR} \\
        \midrule
        \parbox[t]{2mm}{\multirow{5}{*}{\rotatebox[origin=c]{90}{Non-iter.}}} 
        & MSNEA {\footnotesize {\cite{DBLP:conf/kdd/ChenL00WYC22}}} 
        & .260 & .506 & .341 & .334 & .572 & .413 & .332 & .545 & .404 & .612 & .840 & .689 \\
        & EVA {\footnotesize \cite{DBLP:conf/aaai/0001CRC21}} 
        & {.525} & {.827} & {.631} & {.721} & {.918} & {.790} & {.579} & {.809} & {.662} & .775 & .952 & .839 \\
        & MCLEA {\footnotesize {\cite{DBLP:conf/coling/LinZWSW022}}}  
        & {.571} & {.862} & {.675} & {.737} & {.921} & {.803} & {.620} & {.848} & {.704} & {.816} & {.972} & {.874} \\
        & MEAformer {\footnotesize {\cite{chen2023meaformer}}}  
        & \underline{.604} & \underline{.895} & \underline{.708} & \underline{.754} & \underline{.937} & \underline{.818} & \underline{.645} & \underline{.878} & \underline{.729} & \underline{.839} & \underline{.982} & \underline{.892} \\
        & \CC\textbf{UMAEA}  
        & \CC\textbf{.608} & \CC\textbf{.897} & \CC\textbf{.711} & \CC\textbf{.763} & \CC\textbf{.942} & \CC\textbf{.826} & \CC\textbf{.653} & \CC\textbf{.883} & \CC\textbf{.738} & \CC\textbf{.840} & \CC\textbf{.982} & \CC\textbf{.892} \\
        \midrule
        \parbox[t]{2mm}{\multirow{5}{*}{\rotatebox[origin=c]{90}{Iter.}}}
        & MSNEA {\footnotesize {\cite{DBLP:conf/kdd/ChenL00WYC22}}} 
        & .294 & .580 & .391 & .385 & .621 & .463 & .417 & .655 & .500 & .657 & .864 & .726 \\
        & EVA {\footnotesize \cite{DBLP:conf/aaai/0001CRC21}} 
        & {.602} & {.873} & {.699} & {.770} & {.936} & {.829} & {.658} & {.861} & {.734} & .848 & .980 & .899 \\
        & MCLEA {\footnotesize {\cite{DBLP:conf/coling/LinZWSW022}}}  
        & {.646} & {.899} & {.739} & {.790} & {.946} & {.846} & {.696} & {.896} & {.772} & {.881} & {.984} & {.922} \\
        & MEAformer {\footnotesize {\cite{chen2023meaformer}}}  
        & \underline{.656} & \underline{.916} & \underline{.749} & \underline{.793} & \underline{.950} & \underline{.848} & \underline{.703} & \underline{.889} & \underline{.772} & \textbf{.884} & \underline{.988} & \underline{.923} \\
        & \CC\textbf{UMAEA}  
        & \CC\textbf{.670} & \CC\textbf{.921} & \CC\textbf{.763} & \CC\textbf{.801} & \CC\textbf{958} & \CC\textbf{.857} & \CC\textbf{.715} & \CC\textbf{.910} & \CC\textbf{.789} & \CC{.882} & \CC\textbf{.993} & \CC\textbf{.925} \\
        \bottomrule
    \end{tabular}
    }
    \label{tab:appoea}
    \vspace{-0.8cm}
\end{table}

\vspace{-0.3cm}
\subsubsection{\textbf{Baseline Analysis.}}
We attribute the lower performance of translation based methods (e.g., MSNEA) to their reliance on semantics assumptions, which limits their ability to capture the complex structural information among entities for alignment.

Some works \cite{DBLP:conf/ijcai/WuLF0Y019,DBLP:conf/cikm/YangWZQWHH21} hold that 
the structural information plays an important role in the EA task. 
By performing graph convolution over an entity’s neighbors, GCNs can incorporate more structural characteristics of knowledge graphs, while the translation assumption in translation-based models focuses more on the relationship among heads, tails and relations.

\subsection{Model Details}
We reproduce EVA \cite{DBLP:conf/aaai/0001CRC21}, MSNEA \cite{DBLP:conf/kdd/ChenL00WYC22}, MCLEA \cite{DBLP:conf/coling/LinZWSW022} and MEAformer \cite{chen2023meaformer} based on their source code \footnote{\color{blue} \url{https://github.com/cambridgeltl/eva}}$^{,}$\footnote{{\color{blue} \url{https://github.com/lzxlin/MCLEA}}}$^{,}$\footnote{\color{blue} \url{https://github.com/liyichen-cly/MSNEA}}$^{,}$\footnote{\color{blue} \url{https://github.com/zjukg/MEAformer}} with their
original model pipelines unchanged but unifying hyper-parameters.
 Yuan et al. \cite{yuan2023multi} consider the inter-modal effects and mitigate the impact of weak modalities, while Hama et al. \cite{DBLP:journals/access/HamaM23} quantify the importance of modality by embedding the entities into the probability distribution.
 Guo et al. \cite{DBLP:journals/corr/abs-2305-14651} propose the GEEA framework with the mutual variational autoencoder (M-VAE) to mutually encode/decode entities between source and target KGs for both entity alignment and
entity synthesis.
Given that their methods have different goals than ours and were recently published, we did not perform direct comparisons with them in our experiments.

\subsection{Metric Details}
\subsubsection{\textbf{Hits$@$N}} describes the fraction of true aligned 
 target entities that appear in the first N entities of the sorted rank list:
\begin{equation}
  \vspace{-2pt}
    \operatorname{Hits} @ \text{N}=\frac{1}{|\mathcal{S}_{te}|} \sum_{i=1}^{|\mathcal{S}_{te}|} \mathbb{I}[{\text {rank}_i} \leqslant \text{N}]\, ,
\end{equation}
where  ${{\text{rank}}_{i}}$ refers to the rank position of the first correct mapping for the i-th query entities and $\mathbb{I}=1$ if ${\text {rank}_i} \leqslant N$ and 0 otherwise.
$\mathcal{S}_{te}$ refers to the testing alignment set.
\subsubsection{\textbf{MRR}} (Mean Reciprocal Ranking $\uparrow$) is a statistic measure for evaluating many algorithms that produces a list of possible responses to a sample of queries, ordered by probability of correctness. 
In the field of EA, the reciprocal rank of a query entity (i.e., an entity from the source KG) response is the multiplicative inverse of the rank of the first correct alignment entity in the target KG.
MRR is the average of the reciprocal ranks of results for a sample of candidate alignment entities:
\begin{equation}
    \vspace{-2pt}
    \mathbf{MRR}=\frac{1}{|\mathcal{S}_{te}|} \sum_{i=1}^{|\mathcal{S}_{te}|} \frac{1}{\text {rank}_i} \,.
\end{equation}
\subsubsection{\textbf{MR}} (Mean Rank $\downarrow$) computes the arithmetic mean over all individual ranks which is similar to MRR:
\begin{equation}
    \mathbf{MR}=\frac{1}{|\mathcal{S}_{te}|} \sum_{i=1}^{|\mathcal{S}_{te}|} {\text {rank}_i} \,. 
\end{equation}
Note that MR is sensitive to any model performance changes, not only changes that occur below a certain cutoff and therefore reflects the average performance.

\subsection{Future Work \& Discussion}
Knowledge Graphs (KGs) have been empirically validated to provide substantial benefits in a multitude of downstream applications. They serve as significant sources of knowledge supplementation and data augmentation for diverse tasks including, but not limited to, Question Answering \cite{DBLP:conf/semweb/0007CGPYC21,DBLP:conf/jist/0007HCGFP0Z22}, Zero-shot Learning \cite{DBLP:journals/pieee/ChenGCPHZHC23,DBLP:conf/ijcai/ChenG0HPC21,chen2023duet,DBLP:conf/www/GengC0PYYJC21}, and AI4Science \cite{fang2023knowledge,DBLP:conf/aaai/FangZYZD0Q0FC22}.

Despite these advancements, the application of Multi-modal Knowledge Graphs (MMKGs) to such tasks remains relatively unexplored. One plausible reason for this gap is the inherent uncertainty, ambiguity, and occasional missing phenomena associated with various modalities in MMKGs, a challenge particularly prominent within the visual modality, as examined in this paper.

Our objective with this research is to stimulate further academic discourse and exploration in the direction of Multi-modal Entity Alignment (MMEA). We anticipate more scholarly endeavors focusing on MMKG-driven downstream tasks, and we eagerly look forward to the comprehensive understanding and exploitation of the untapped potential of multi-modal KGs within the Semantic Web community.

Moreover, there remain opportunities for future research related to this work, such as evaluating our techniques in the context of incompleteness in other modalities (e.g., attribute), and investigating effective techniques to  utilize more detailed visual contents for MMEA. There is  potential for enhancing UMAEA’s efficiency, we also view this as a direction for future research which has not been explored in depth.

\end{document}